\documentclass[10pt,a4paper,twocolumn]{article}
\usepackage[a4paper,margin=26mm,bottom=27mm,includefoot,footskip=18pt,columnsep=18pt]{geometry}
\usepackage[T1]{fontenc}
\usepackage[utf8]{inputenc}
\usepackage{newtxtext,newtxmath}
\usepackage{amsmath,booktabs,tabularx,graphicx,xcolor,etoolbox}
\usepackage{multirow,placeins,capt-of,float,adjustbox,xurl}
\usepackage{setspace,colortbl,arydshln}
\usepackage[numbers,sort&compress]{natbib}
\usepackage[font=normalsize,labelfont=bf]{caption}
\usepackage{dblfloatfix}
\usepackage[hidelinks]{hyperref}
\usepackage[protrusion=false]{microtype}
\renewcommand{\small}{\fontsize{10}{12}\selectfont}
\renewcommand{\footnotesize}{\fontsize{10}{12}\selectfont}

\definecolor{paperblue}{HTML}{EDF4FA}
\definecolor{papergreen}{HTML}{EDF6F2}
\definecolor{paperpurple}{HTML}{F3F0F8}
\newcommand{\yes}{\textcolor[HTML]{377F72}{\ensuremath{\checkmark}}}
\newcommand{\partly}{\textcolor[HTML]{967032}{\ensuremath{\triangle}}}
\newcommand{\none}{\textcolor{black!45}{---}}
\newcolumntype{C}[1]{>{\centering\arraybackslash\hspace{0pt}}m{#1}}

\newcolumntype{Y}{>{\centering\arraybackslash\hspace{0pt}}X}

\newcommand{\modeN}{\texttt{new\_session}}
\newcommand{\modeS}{\texttt{same\_session}}
\hypersetup{pdftitle={AstroMemBench: A Benchmark for Memory Systems in Embodied Spacecraft Agents},
  pdfauthor={You Wu, Guhong Zhang, Hongwei Li, Xinyue Hong, Yuan Jia},
  pdfsubject={Agent memory evaluation for embodied spacecraft agents},pdfcreator={you wu}}

\newcommand{\PaperKeywords}{Agent memory; Embodied spacecraft agent; Memory benchmark; Spacecraft autonomy; History-dependent task}

\title{AstroMemBench: A Benchmark for Memory Systems\\in Embodied Spacecraft Agents}
\author{\normalsize You Wu\textsuperscript{1}, Guhong Zhang\textsuperscript{1,*}, Hongwei Li\textsuperscript{2}, Xinyue Hong\textsuperscript{1}, Yuan Jia\textsuperscript{3}\\[6pt]
\normalsize\textsuperscript{1}School of Aeronautics and Astronautics, Zhejiang University\\
\normalsize\textsuperscript{2}College of Information Science and Electronic Engineering, Zhejiang University\\
\normalsize\textsuperscript{3}Institute of Advanced Technology, Zhejiang University\\
\normalsize Hangzhou 310027, China\\[3pt]
\normalsize\textsuperscript{*}Corresponding author: \href{mailto:ghzhang@zju.edu.cn}{ghzhang@zju.edu.cn}}
\date{}

\begin{document}
\raggedbottom
\twocolumn[\begin{@twocolumnfalse}
\maketitle
\begin{abstract}
Long-term spacecraft autonomy requires embodied agents to use operational history under communication delays, time-critical operations, environmental uncertainty, complex physical interactions, power and computational constraints, and safety and reliability requirements. We introduce AstroMemBench and a Scenario--Challenge--Capability design method that maps these requirements to executable memory tests. The benchmark covers 16 memory subchallenges through 140 templates and 2240 instances across nine tasks in planetary surface operations, on-orbit servicing, and deep-space exploration. A deterministic environment scores actions for task completion, state changes, safety, and resource use. Ablations under nine conditions show that the benchmark is sensitive to how accurately agents recover and use historical information and to differences in memory-system design. Main experiments compare six configurations. Mem0, Chain-of-Memory, LightMem, and retrieval-augmented generation (RAG) improve same-session mean scores over native Pi Agent, with no clear advantage of the three memory systems over RAG. After a session restart, Mem0 outperforms RAG by 12.25 points. Benefits favor instruction retention and result reuse over ongoing state tracking and energy reservation. AstroMemBench supports the design, development, and selection of memory systems for embodied spacecraft agents. AstroMemBench 1.0 and its evaluation code are publicly available: \url{https://github.com/flyingView/AstroMemBench-v1}.

\end{abstract}
\vspace{14pt}
\end{@twocolumnfalse}]
\section{Introduction}

Advances in planetary surface operations~\cite{hirano2026soraq}, on-orbit servicing~\cite{rybus2024manipulators}, and deep-space exploration~\cite{zhang2026smallbodies} increase demand for unattended, long-term autonomous spacecraft operation~\cite{castano2022operations,esa2025technology2040}. Agents based on large language models and their multimodal extensions support such autonomy through perception, reasoning, and tool use; related studies in simulated and experimental orbital operations and on-orbit servicing have explored closed-loop execution and experience reuse~\cite{carrasco2025vlmoperators,wu2026spacemind,carrasco2026guide}. Within these agents, memory systems preserve, organize, and update interaction records so that facts and experience remain available for decisions beyond a finite context window~\cite{chhikara2025mem0,xu2026com}. For embodied spacecraft agents on long missions, this ability supports continued adherence to task requirements, resumption from the correct execution state, and effective reuse of prior experience.

\begin{figure*}[tp]
\centering\includegraphics[width=\textwidth]{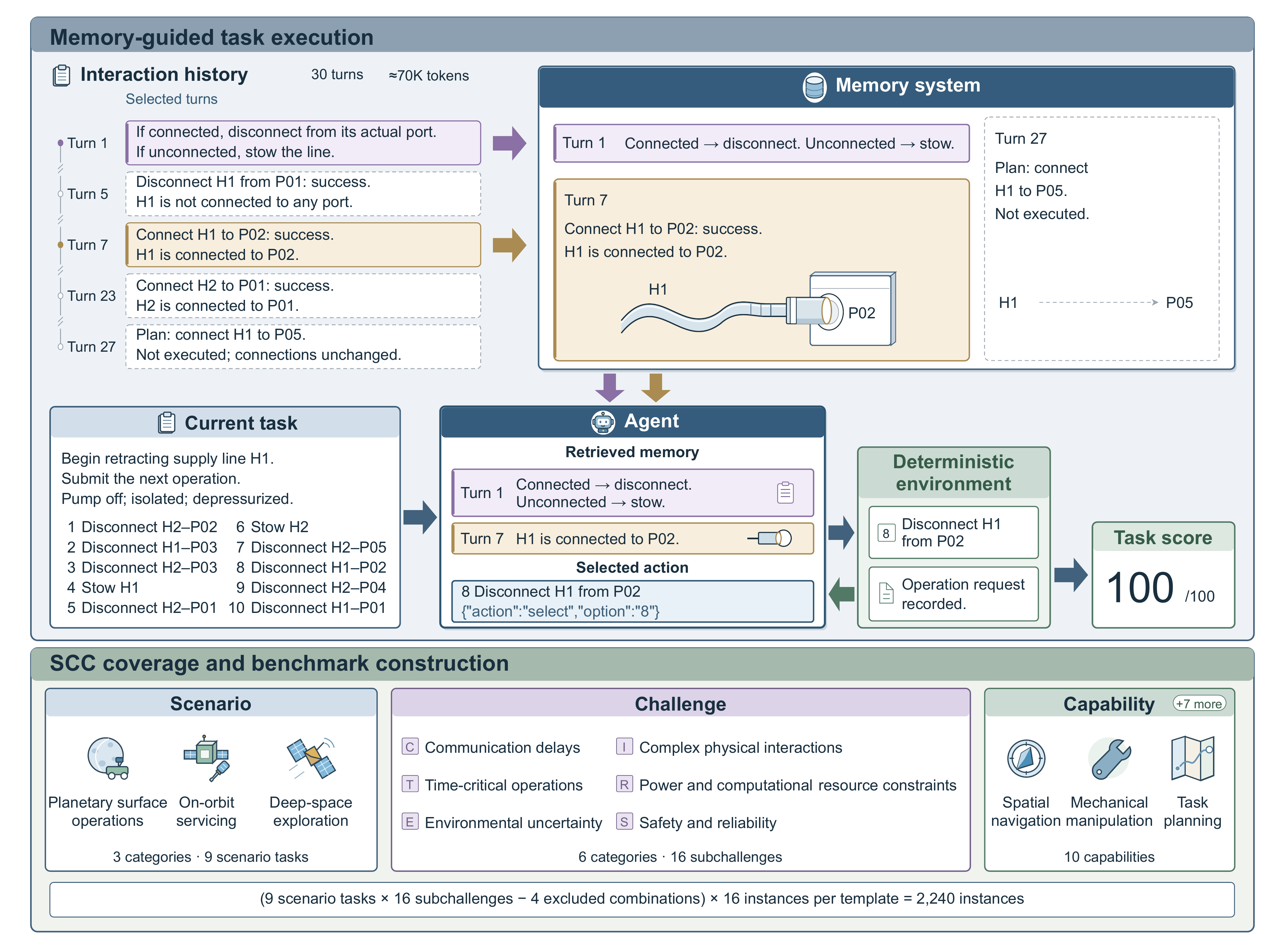}
\caption{Overview of AstroMemBench. Top: a schematic supply-line task with a simplified action menu shows how historical evidence supports action selection through memory, with deterministic environment scoring. Bottom: scenario, challenge, and capability coverage and test-set construction.}
\label{fig:challenge-examples}
\end{figure*}

However, the suitability of current memory systems for spacecraft tasks requires dedicated evaluation. Mem0, CoM, and LightMem continue to advance~\cite{chhikara2025mem0,xu2026com,fang2026lightmem}, but their ability to meet specialized operational requirements remains to be established. In building information modeling, for example, general-purpose systems retrieved relevant content yet omitted required domain information~\cite{du2026ifcmemorybench}. This need for domain-specific validation is not fully addressed by existing memory benchmarks. MemoryAgentBench notes that static long-context evaluations inadequately capture agents' incremental information accumulation and updating; recent research also identifies gaps between benchmark scale, metrics, and practical needs~\cite{hu2026memoryagentbench,jiang2026anatomy}. The main experiments on SimpleMem~\cite{liu2026simplemem} and LightMem~\cite{fang2026lightmem}, two recent memory systems, still use the earlier conversational benchmarks LoCoMo~\cite{maharana2024locomo} and LongMemEval~\cite{wu2025longmemeval}. These benchmarks provide a common basis for testing long-term conversational memory, whereas evaluations of spacecraft agents must also test whether historical information helps them act correctly under the operational constraints of space missions. A dedicated benchmark is therefore needed to guide memory system design and selection.

Remote operation and limited intervention place particular demands on the use of history in space missions. The European Space Agency's \emph{Technology 2040} identifies six challenges for autonomous space systems: communication delays, time-critical operations, environmental uncertainty, complex physical interactions, power and computational resource constraints, and safety and reliability~\cite{esa2025technology2040}. Communication delays hinder timely ground confirmation, requiring agents to follow existing requirements and distinguish valid instructions from superseded ones; time-critical operations require agents to use valid information within action windows and remember pending tasks. Unknown environments and complex interactions require agents to revise judgments using prior observations, actual operation results, and environmental changes, while distinguishing plans from completed actions. Limited energy and computation also make experience and result reuse directly relevant to task cost. Safety imposes further demands: for example, deep-space flybys cannot pause when anomalies occur, and delayed responses may cause mission failure~\cite{castano2022operations}, making valid safety checks and correct task resumption after faults particularly important. To assess these requirements, memory evaluation must link history to action consequences, including task completion, safety constraints, and resource use.

We therefore develop \textbf{AstroMemBench, a benchmark for memory systems in embodied spacecraft agents} (Fig.~\ref{fig:challenge-examples}), translating these operational requirements into executable, history-dependent tasks to examine the role of memory. Our main contributions are:

\begin{enumerate}
\item \textbf{We introduce the three-dimensional SCC design method for domain coverage.} Scenario--Challenge--Capability (SCC) organizes benchmark construction through a scenario--challenge applicability matrix, capability associations, and execution checks, translating domain requirements into inspectable, revisable test coverage.
\item \textbf{We construct a dedicated memory test set for embodied spacecraft agents.} We refine six operational challenge categories into 16 memory subchallenges across nine tasks in planetary surface operations, on-orbit servicing, and deep-space exploration, yielding 140 templates and 2240 instances. Each instance establishes task context through 30 historical interaction turns; subsequent execution is scored deterministically by state changes, safety conditions, and resource consequences.
\item \textbf{We demonstrate the benchmark's sensitivity to history availability and differences among memory systems.} Nine ablation conditions test the effects of historical background, session boundaries, and context windows. Comparisons of scores, baseline-relative gains, and operational costs under same-session and new-session conditions, with each instance isolated, reveal how memory benefits vary across systems and task requirements.
\end{enumerate}

\section{Related Work}

\subsection{Agent Memory Systems}
Agent memory systems support decisions over time by retaining, updating, and retrieving historical information. RAG retrieves external fragments to support generation~\cite{lewis2020rag}; Mem0 also extracts and integrates important interaction content~\cite{chhikara2025mem0}. CoM preserves raw interaction nodes and assembles evidence chains after retrieval, whereas LightMem uses compression, staged processing, and offline updating to reduce maintenance overhead~\cite{xu2026com,fang2026lightmem}. Given these different approaches to history reuse, domain-specific evaluation must establish whether each preserves valid instructions, equipment states, and safety conditions and uses them correctly during spacecraft tasks. We compare RAG, Mem0, CoM, and LightMem under common task conditions to assess both performance and model-call overhead.

\subsection{Benchmarks for Memory System Evaluation}
Early memory benchmarks focused on retaining and using information over long interactions. LoCoMo tests long-term conversational memory through multisession question answering, event summarization, and generation~\cite{maharana2024locomo}, while LongMemEval distinguishes cross-session reasoning, knowledge updating, and temporal reasoning~\cite{wu2025longmemeval}. MemoryAgentBench uses incremental interaction to test retrieval, test-time learning, long-range understanding, and outdated-information handling~\cite{hu2026memoryagentbench}. Recent work broadens both what memory retains and how it is used: Mem-Gallery tests retention, reasoning, and management of visual and textual information over long interactions; LoCoMo-Plus tests whether subsequent answers correctly apply states, goals, and implicit constraints that are not directly queried~\cite{memgallery,li2026locomoplus}. Together, these benchmarks extend evaluation from factual recall to temporal reasoning, state updating, and implicit constraints. Their emphasis on conversational answers and offline outputs leaves the consequences of memory use during spacecraft task execution largely untested.

Task-execution studies increasingly connect memory to agent behavior. AMA-Bench constructs memory questions from trajectories containing states, actions, observations, and tool outputs, extending evaluation from conversations to agent experiences~\cite{amabench}. Mem2ActBench tests whether historical constraints support tool-call argument generation. PM-Bench tests delayed intentions triggered by temporal or event cues; MemoryArena uses interdependent cross-session subtasks to examine how prior experience affects later decisions~\cite{shen2026mem2actbench,pmbench,he2026memoryarena}. Alongside task performance, Wolff and Bennati~\cite{wolff2026costmemory} examine communication, computation, and storage overhead in a simulated cloud--edge network, although their accuracy comparisons still use tasks from general conversational benchmarks. These studies move beyond conversational recall toward tool use, interactive execution, and memory costs. Their task environments remain largely general-purpose, however, without systematic coverage of the operational constraints that distinguish spacecraft tasks.
\begin{table*}[tp]
\caption{Capability coverage of representative benchmarks. Five general capabilities and six space-relevant capabilities derived from our six operational challenges are shown; space-relevant capabilities may also be tested in other domains.}
\label{tab:benchmarks}
\setstretch{1}
\centering\fontsize{10}{12}\selectfont
\setlength{\tabcolsep}{1.1pt}\renewcommand{\arraystretch}{1.12}
\begin{tabular*}{\textwidth}{@{\extracolsep{\fill}}l*{11}{c}@{}}
\toprule
 & \multicolumn{5}{c}{\textbf{General capabilities}} & \multicolumn{6}{c}{\textbf{Space-relevant capabilities}}\\
\cmidrule(lr){2-6}\cmidrule(l){7-12}
\textbf{Benchmark} & \shortstack{Factual\\recall} & \shortstack{Time\\assoc.} & \shortstack{State\\maint.} & \shortstack{Visual\\memory} & \shortstack{Exper.\\transfer} & \shortstack{Instr.\\validity} & \shortstack{Action\\window} & \shortstack{Unk.\\environ.} & \shortstack{Phys.\\interact.} & \shortstack{Res.\\cost} & \shortstack{Safety\\and rel.}\\
\midrule
LoCoMo~\cite{maharana2024locomo} & \yes & \yes & \yes & \yes & \partly & \none & \none & \none & \none & \none & \none\\
LongMemEval~\cite{wu2025longmemeval} & \yes & \yes & \yes & \none & \partly & \none & \none & \none & \none & \none & \none\\
MemoryAgentBench~\cite{hu2026memoryagentbench} & \yes & \yes & \yes & \none & \yes & \none & \none & \none & \none & \none & \none\\
Mem-Gallery~\cite{memgallery} & \yes & \yes & \yes & \yes & \yes & \none & \none & \none & \none & \none & \none\\
LoCoMo-Plus~\cite{li2026locomoplus} & \yes & \yes & \yes & \none & \partly & \partly & \none & \none & \none & \none & \none\\
\hdashline[2pt/2pt]
AMA-Bench~\cite{amabench} & \yes & \yes & \yes & \none & \none & \none & \none & \partly & \none & \none & \none\\
Mem2ActBench~\cite{shen2026mem2actbench} & \yes & \yes & \yes & \none & \partly & \yes & \none & \none & \none & \none & \none\\
PM-Bench~\cite{pmbench} & \yes & \yes & \yes & \none & \none & \yes & \yes & \partly & \none & \none & \none\\
MemoryArena~\cite{he2026memoryarena} & \yes & \yes & \yes & \none & \yes & \yes & \partly & \partly & \none & \partly & \none\\
\hdashline[2pt/2pt]
FindingDory~\cite{findingdory} & \yes & \yes & \yes & \yes & \none & \none & \none & \yes & \partly & \partly & \none\\
LMEE-Bench~\cite{wang2026lmee} & \yes & \partly & \yes & \yes & \partly & \none & \none & \yes & \partly & \partly & \none\\
MIKASA-Robo~\cite{mikasa} & \yes & \yes & \yes & \yes & \partly & \none & \partly & \yes & \yes & \none & \none\\
RoboMME~\cite{dai2026robomme} & \yes & \yes & \yes & \yes & \partly & \none & \yes & \yes & \yes & \none & \none\\
AstroAgentBench\textsuperscript{*}~\cite{wang2026astroagentbench} & \partly & \partly & \partly & \none & \yes & \none & \partly & \partly & \none & \yes & \partly\\
\midrule

\textbf{AstroMemBench} & \yes & \yes & \yes & \yes & \partly & \yes & \yes & \yes & \yes & \yes & \yes\\
\bottomrule
\end{tabular*}
\par\vspace{3pt}
\noindent\begin{minipage}{\textwidth}\fontsize{10}{12}\selectfont
\yes: direct test; \partly: related task, procedure, or metric; \none: no corresponding test identified. Marks summarize capability coverage based on representative tasks.
State maintenance includes fact updating and prior-experience applicability. Experience transfer counts known-case matching, preference reuse, and exact demonstration reproduction as partial coverage.
Action windows require temporal or event constraints on actions. Unknown environments include currently unobservable states. Path and monetary costs partially cover resource cost. Safety and reliability include avoiding repeated failures using historical feedback and resuming after obstructions.
\textsuperscript{*}AstroAgentBench is a space-agent benchmark that checks temporal, state, window, and resource constraints in planning outputs; transfer reuses training-task notes. Appendix~\ref{app:coverage-criteria} gives full criteria and source evidence.
\end{minipage}
\end{table*}

In embodied domains relevant to space missions, memory evaluation centers on exploration and manipulation. For exploration, FindingDory tests navigation and revisiting using prior object-interaction records, while LMEE-Bench combines multi-goal navigation with memory question answering to test how exploration history supports later tasks~\cite{findingdory,wang2026lmee}. For manipulation, MIKASA-Robo distinguishes object, spatial, sequential, and capacity memory in tabletop tasks. RoboMME tests temporal, spatial, object, and procedural memory, sometimes imposing operation windows, stage prerequisites, and termination on incorrect actions~\cite{mikasa,dai2026robomme}. These predominantly terrestrial benchmarks test embodied abilities relevant to space missions, but do not jointly cover spacecraft memory requirements arising from communication delays and interruptions, long missions, and limited energy. In particular, instruction validity, equipment states across stages, energy reservation for later tasks, and task resumption after faults remain incompletely covered.

Table~\ref{tab:benchmarks} summarizes these 13 memory benchmarks and includes AstroAgentBench, discussed next, as a related domain reference. Horizontal rules separate general memory, task execution, and domain tasks. Across these groups, coverage progresses from conversational recall to task execution and embodied interaction, while spacecraft requirements remain distributed across separate evaluations. AstroMemBench brings the six operational challenge categories into a common set of spacecraft tasks and evaluates memory through actions and their state, resource, and safety consequences. Capability criteria and broader protocol comparisons appear in Appendix~\ref{app:related-work}.

\subsection{Space Agents and Memory Evaluation Requirements}
Space agents combine language-model reasoning with environmental feedback and action. Carrasco et al.\ generate orbital operation commands from KSPDG interface images, exploring vision--language-driven closed-loop control~\cite{carrasco2025vlmoperators}. Experience reuse is central to two further approaches: in orbital interception simulations, GUIDE extracts and updates state-conditioned decision rules from execution trajectories, improving policies across episodes without changing model weights~\cite{carrasco2026guide}; for on-orbit servicing, SpaceMind combines visual perception, reasoning, and tools with hierarchical memory and skill accumulation~\cite{wu2026spacemind}. These studies demonstrate the feasibility of closed-loop execution and experience accumulation in space tasks. Extending such operation over long periods requires dedicated tests of how memory supports instruction validity, state maintenance across stages, and safe execution.

AstroAgentBench covers space-related scheduling, observation planning, and constellation design, primarily evaluating agents that assist ground-based planning and design~\cite{wang2026astroagentbench}. It is designed to assess the overall capabilities of these planning assistants and does not specifically target their memory systems. AstroMemBench targets memory systems in embodied spacecraft agents and compares how different memory mechanisms support embodied execution on the same tasks with the same interaction histories.

\section{AstroMemBench and Its Design Method}
\label{sec:benchmark}

AstroMemBench tests whether memory helps embodied spacecraft agents act correctly and efficiently using past experience. Each instance pairs historical interactions with a subsequent task: after reading the history, the agent acts through tools in a deterministic environment, which scores task completion, state changes, safety conditions, and resource consequences.

\subsection{The SCC Benchmark Design Method}
General-purpose benchmarks test information retention, updating, and access; domain-specific benchmarks must further establish whether these capabilities meet specialized operational requirements. To make this additional coverage explicit, our three-dimensional Scenario--Challenge--Capability (SCC) design method translates requirements inadequately covered by general tests into representative tasks. Associations across the three dimensions allow designers to inspect how these tasks complement general evaluation.

The challenge dimension identifies memory requirements that need particular attention: domain-specific issues and common issues subject to stronger constraints or more serious consequences in that domain. Benchmark designers should identify these difficulties as fully as possible and specify what history each challenge requires, when the agent needs it, and what happens if it is omitted or misused. This translates domain requirements into measurable memory tasks.

The scenario dimension defines the application scope. Because the same memory challenge may involve different evidence, operating conditions, and consequences across tasks, benchmark designers need to specify the main scenarios and task types in advance. This scope guides coverage of operational stages and situations at a feasible scale and provides a common basis for expanding tasks and refining systems.

The capability dimension connects domain task requirements to agent execution. Scenarios and challenges describe task contexts and difficulties, but agents act through modular capabilities. General-purpose agents call tools, execute code, or schedule tasks; domain agents also need foundational capabilities such as navigation, mechanical manipulation, and equipment configuration. Memory must supply applicable rules, parameters, states, and experience for these capabilities. A benchmark should therefore list the required capabilities and test whether history supports their correct use and coordination.

Construction begins by identifying challenges, scenarios, and capabilities and estimating the benchmark's scale from their granularity, applicable combinations, and instances per template. We then plan templates in this three-dimensional space, linking each to a scenario task, a memory subchallenge, and one or more required capabilities. To achieve sufficient, balanced coverage, we assign tasks to applicable combinations, check for concentration in a few scenarios or challenges, and fill capability gaps. Each template then specifies decisive history, the current task, available actions, and scoring consequences. History reconstruction and execution replay test whether these specifications yield measurable tasks; any coverage gaps or execution inconsistencies lead back to template revision.

\subsection{Memory Requirements and Task Coverage}
We refine the six operational challenges in the European Space Agency's \emph{Technology 2040}~\cite{esa2025technology2040} into 16 memory subchallenges (Table~\ref{tab:challenges}), specifying what historical evidence is needed, when it is needed, and what happens if it is misused. Under communication delays, for example, agents must follow existing instructions after contact is lost, resolve out-of-order instructions by issuance relationships, and confirm delivery through reception acknowledgments. These requirements concern valid instructions, version relationships, and delivery status, corresponding to C1, C2, and C3. The remaining challenges similarly define observable history dependencies, bringing temporal, environmental, physical-interaction, resource, and safety requirements into instance construction and scoring.

\begin{table*}[tp]
\caption{Typical test construction for the 16 memory subchallenges}
\label{tab:challenges}
\setstretch{1}
\centering\small
\setlength{\tabcolsep}{4pt}\renewcommand{\arraystretch}{1.08}
\begin{tabularx}{\textwidth}{@{}C{40mm}Y@{}}
\toprule[.9pt]
\textbf{Subchallenge} & \textbf{Construction of the memory test}\\
\midrule[.65pt]
C1 Instruction retention & Distribute ground instructions across the history; require continued adherence to still-valid operational requirements after communication is lost.\\
C2 Rule disambiguation & Present issuance, replacement, and revocation records out of order; require action selection based on validity relationships.\\
C3 Downlink confirmation & Mix transmission attempts, failure records, and reception acknowledgments; require determining whether delivery is complete.\\
\midrule[.25pt]
T1 Validity assessment & Provide validity conditions for rules, data, or credentials; require assessing their continued applicability at the current time.\\
T2 Deferred tasks & Register subsequent tasks early in the history; require recalling and executing them when temporal or event triggers occur.\\
\midrule[.25pt]
E1 Experience matching & Provide experience from different environments or operating conditions; require selecting an operational plan applicable to current conditions.\\
E2 Experience revision & Add records of environmental changes after prior experience; require revising current operations accordingly.\\
E3 Localization recovery & Establish observation--location correspondences in the history; require recovering location from current observations without location labels.\\
\midrule[.25pt]
I1 Single-operation state & Distinguish a successful supply-line connection from an unexecuted connection plan; accordingly choose disconnection before stowing or direct stowing.\\
I2 Multi-operation state & Record progress and cumulative physical changes caused by multiple operations; require arranging subsequent operations based on the current state.\\
\midrule[.25pt]
R1 Energy-use experience & Provide historical energy-use experience for routes or operations; require selecting a plan that satisfies energy constraints.\\
R2 Energy reservation & Register the energy needs of subsequent tasks in advance; require reserving the corresponding energy during current operations.\\
R3 Result reuse & Provide existing computed results matching current conditions; include simulated energy consumed by unnecessary recomputation in scoring.\\
\midrule[.25pt]
S1 Safety prerequisites & Include safety records for mismatched objects or records that are no longer valid; require confirming valid safety evidence before operations.\\
S2 Capability updates & Record functional changes caused by faults, repairs, or equipment replacement; require selecting operations according to updated equipment capabilities.\\
S3 Fault recovery & Record progress before a fault and actual feedback after recovery; require resuming the task from the correct stage.\\
\bottomrule[.9pt]
\end{tabularx}
\par\vspace{3pt}
\noindent\begin{minipage}{\textwidth}\footnotesize
The 140 templates instantiate these subchallenges across nine scenario tasks.
C: communication delays; T: time-critical operations; E: environmental uncertainty;
I: complex physical interactions; R: power and computational resource constraints;
S: safety and reliability.
\end{minipage}
\end{table*}

Three scenario categories each contain three tasks: planetary surface operations include area surveying, sampling at designated sites, and cargo transport; on-orbit servicing includes target inspection, equipment repair, and on-orbit replenishment; deep-space exploration includes cruise support, scientific observation, and data return. Crossing each task with 16 subchallenges yields 16 candidate templates per task and 144 combinations across nine tasks. Data return has little need for localization recovery, single-operation state, or multi-operation state; target inspection is largely unaffected by accumulated physical effects of multiple operations. Excluding these combinations leaves 15 templates for target inspection, 13 for data return, and 16 for each remaining task, totaling 140. Appendix~\ref{app:scc} explains the exclusions.

\begin{figure*}[tp]
\centering\includegraphics[width=\textwidth]{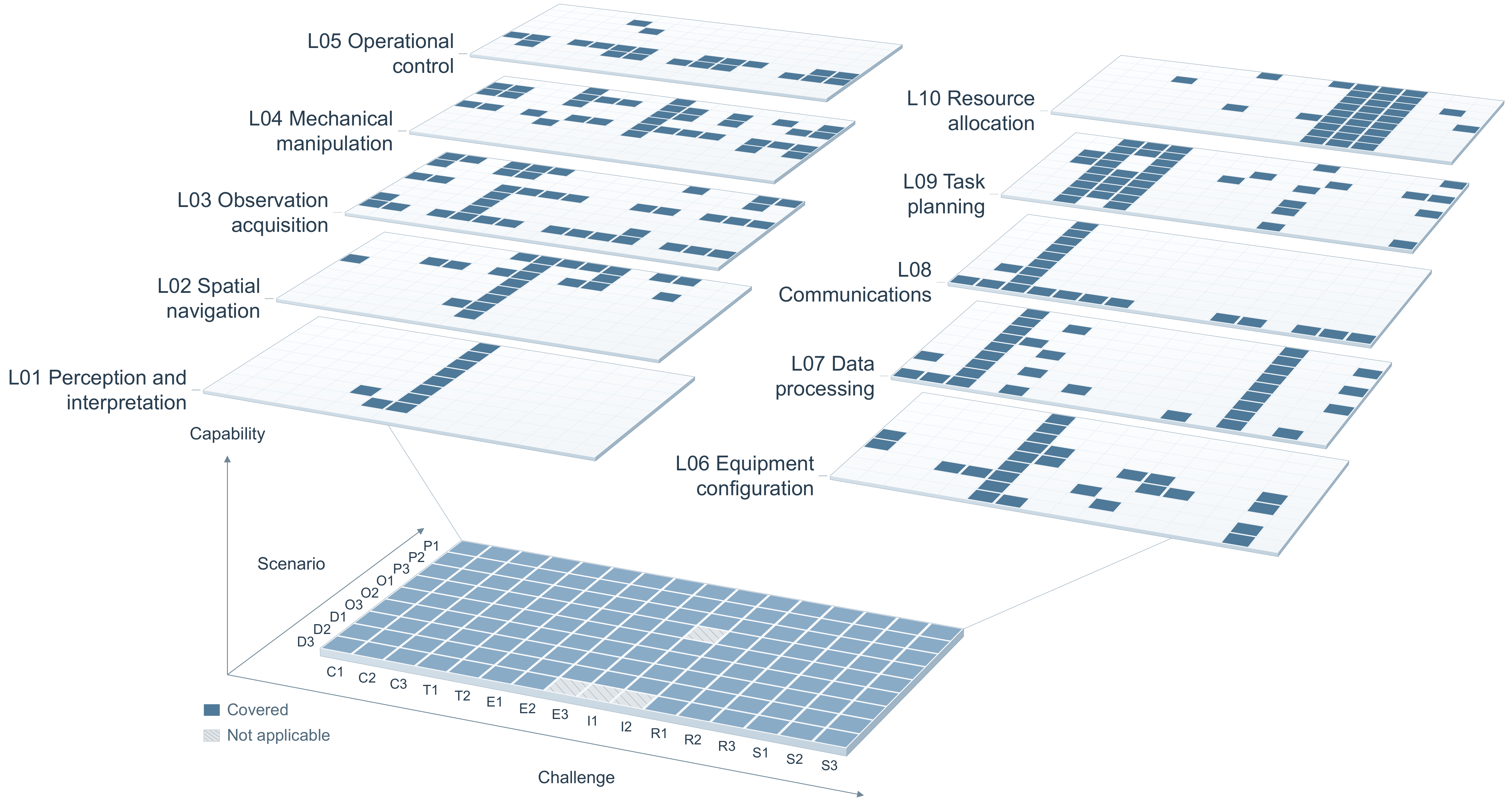}
\caption{Coverage under the three-dimensional SCC design method. The base crosses nine scenario tasks with 16 subchallenges. Each applicable cell defines a template testing that subchallenge in that scenario task: 140 templates cover all applicable combinations; four inapplicable combinations are hatched. The ten capability layers share these coordinates, with colored cells marking capabilities used by each template (285 associations). Two staggered columns expose every layer's projection. Each capability spans multiple tasks and subchallenges.}
\label{fig:associations}
\end{figure*}

The ten foundational capabilities are perception and interpretation, spatial navigation, observation acquisition, mechanical manipulation, operational control, equipment configuration, data processing, communications, task planning, and resource allocation. Each of the 140 templates uses one to five capabilities, yielding 285 template--capability associations. For example, the supply-line template pairs on-orbit replenishment with the single-operation state subchallenge and requires mechanical manipulation to disconnect or stow a line according to its historical connection state. Figure~\ref{fig:associations} shows the resulting coverage. Capability labels describe which abilities the templates cover; execution consequences determine instance scores.

\subsection{Template Instantiation and Test Instance Organization}
We construct the test set through template design, instance expansion, and validation. Instances derived from the templates summarized in Table~\ref{tab:challenges} vary in historical events and task conditions while preserving each template's core memory requirement. Paired instances can share the same current request yet require different actions because their histories establish different states. Expert review and automated checks verify consistency among historical evidence, environment states, and scoring rules.

Each template yields 16 instances: 6 easy, 6 medium, and 4 hard. While the core memory requirements and assessment criteria remain fixed, difficulty varies with object and record counts, evidence dispersion, and rule updates, revocations, or state reversals. For equipment-repair parameter reuse, easy instances place calibration information and results in one record; medium instances separate computation initiation and completion; hard instances further disperse equipment configuration, operating conditions, and results, requiring links through work orders. These variations in task rules, events, maps, and resource values are implemented in an instance-specific deterministic environment.

Each instance contains past interaction history, the current initial environment state, an initial public observation, action-dependent state-transition rules, and a scoring function. History records instructions, objects, operations, and actual feedback, while the initial observation provides the current request and interfaces. As the agent submits tool actions, the environment updates its state under template rules, returns feedback, and assigns a score of 0--100 at termination. A task may end after a single choice or require multistep execution, so action turns and historical turns are specified separately.

Supply-line instances FM-I1-O3-001 and 002 have identical current requests and operation menus: begin retracting H1 and submit the next operation. Only historical turn 7 differs: the former records successful connection of H1 to P02; the latter records an unexecuted connection plan. The correct actions are therefore to disconnect H1 from P02 and to stow H1 directly, respectively. This pair isolates the effect of an actual historical event on action choice, testing memory-supported state judgment.

History also affects task cost. In equipment-repair parameter-reuse instance FM-R3-O2-001, history contains a computed result matching the current equipment, operation, configuration, and conditions. Direct reuse earns 100 points; correct submission after one computation-service call earns 90; incorrect submission earns 0. The service consumes 30 units of simulated computation energy from an initial budget of 300. Scoring thus captures both action correctness and effective result reuse. Figure~\ref{fig:challenge-examples} shows evaluation, SCC construction, and history-dependency examples; Appendix~\ref{app:protocol-details} gives scoring rules.

Each instance supplies 30 read-only historical turns sequentially, allowing the evaluated system to preserve and organize information through its own mechanisms. History has three components with distinct roles: target evidence determines current rules, states, or results; task-related distractors, including old versions, other objects, unexecuted plans, and failure records, require applicability judgments; routine background controls information load. Histories with background distribute complete operational paragraphs across 30 turns, with approximately 70K input tokens. To isolate the effect of this load, clean histories remove only routine background, retaining target evidence, distractors, images, clocks, and turn boundaries, with identical initial states and scoring criteria. Visual localization uses frozen NASA images without location answers or auxiliary captions. Appendix~\ref{app:scc} details historical materials, paired instances, and environment implementation.

\subsection{Test-Set Scale and Quality Validation}
The 140 templates yield 2240 instances across nine scenario tasks: 840 easy, 840 medium, and 560 hard. Table~\ref{tab:corpus} summarizes their distribution by task.

\begin{table*}[tbp]\centering
\caption{Scenario tasks, templates, and instances in AstroMemBench}
\label{tab:corpus}
\setstretch{1}
\centering
\newlength{\corpusTplWidth}\newlength{\corpusInstWidth}
\newlength{\corpusSmallWidth}\newlength{\corpusMediumWidth}
\ifdim\linewidth<\textwidth
  \small\setlength{\tabcolsep}{0.6pt}
  \setlength{\corpusTplWidth}{16pt}\setlength{\corpusInstWidth}{19pt}
  \setlength{\corpusSmallWidth}{18pt}\setlength{\corpusMediumWidth}{31pt}
  \def\corpusTplLabel{Tpl.}\def\corpusInstLabel{Inst.}
  \def\corpusGroupLabel{Per template}
  \def\corpusAbbreviationNote{Tpl.: templates; Inst.: total instances. }
\else
  \small\setlength{\tabcolsep}{1pt}
  \setlength{\corpusTplWidth}{4.5em}\setlength{\corpusInstWidth}{3.9em}
  \setlength{\corpusSmallWidth}{2.8em}\setlength{\corpusMediumWidth}{3.8em}
  \def\corpusTplLabel{Templates}\def\corpusInstLabel{Total\newline instances}
  \def\corpusGroupLabel{Instances per template}
  \def\corpusAbbreviationNote{}
\fi
\renewcommand{\arraystretch}{1.12}
\begin{tabularx}{\linewidth}{@{}Y C{\corpusTplWidth}C{\corpusSmallWidth}C{\corpusMediumWidth}C{\corpusSmallWidth}C{\corpusInstWidth}@{}}\toprule
\multirow[c]{2}{=}{\centering\textbf{Scenario task}} & \multirow[c]{2}{=}{\centering\textbf{\corpusTplLabel}} & \multicolumn{3}{c}{\textbf{\corpusGroupLabel}} & \multirow[c]{2}{=}{\centering\textbf{\corpusInstLabel}}\\
\cmidrule(lr){3-5}
 & & Easy & Medium & Hard & \\\midrule
\multicolumn{6}{c}{\textbf{P Planetary surface spacecraft}}\\
P1 Area surveying & 16 & 6 & 6 & 4 & 256\\
P2 Sampling at designated sites & 16 & 6 & 6 & 4 & 256\\
P3 Cargo transport & 16 & 6 & 6 & 4 & 256\\
\multicolumn{6}{c}{\textbf{O On-orbit servicing spacecraft}}\\
O1 Target inspection & 15 & 6 & 6 & 4 & 240\\
O2 Equipment repair & 16 & 6 & 6 & 4 & 256\\
O3 On-orbit replenishment & 16 & 6 & 6 & 4 & 256\\
\multicolumn{6}{c}{\textbf{D Deep-space exploration spacecraft}}\\
D1 Cruise support & 16 & 6 & 6 & 4 & 256\\
D2 Scientific observation & 16 & 6 & 6 & 4 & 256\\
D3 Data return & 13 & 6 & 6 & 4 & 208\\\midrule
\textbf{Full test set} & \textbf{140} & $\times$ & \textbf{16} & $=$ & \textbf{2240}\\
\bottomrule\end{tabularx}
\par\vspace{3pt}\noindent\begin{minipage}{\linewidth}\footnotesize
\corpusAbbreviationNote The full test set contains 840 easy, 840 medium, and 560 hard instances. Each template involves 1--5 capabilities; each instance has 30 historical turns, approximately 70K tokens in the version with routine background.
\end{minipage}
\end{table*}

A specified random strategy without history access earns full scores on an average of 1.07 of the 16 instances per template. The template-average probability of full scores on at least 8 instances is 0.0229\% (Table~\ref{tab:guessing-counts}). Random choices thus rarely yield full scores across multiple instances, showing how evaluation across a template's instances limits the influence of chance successes. Appendix~\ref{app:protocol-details} gives the strategy and calculations.

\begin{table*}[tbp]
\caption{Random-guessing probability of full scores on at least $k$ of 16 instances per template. Values summarize the average and maximum across 140 template-specific calculations.}
\label{tab:guessing-counts}
\setstretch{1}
\centering\small
\setlength{\tabcolsep}{3pt}\renewcommand{\arraystretch}{1.10}
\setlength{\aboverulesep}{2pt}\setlength{\belowrulesep}{2pt}
\begin{tabularx}{\textwidth}{@{}C{10mm}Y Y@{}}
\toprule[0.9pt]
\textbf{$k$} & \textbf{Template average (\%)} & \textbf{Template maximum (\%)}\\
\midrule[0.7pt]
1 & 55.6601 & 99.8824\\
4 & 4.8511 & 66.7455\\
8 & 0.0229 & 1.7686\\
16 & $4.24\times10^{-12}$ & $4.67\times10^{-10}$\\
\bottomrule[0.9pt]
\end{tabularx}
\end{table*}

Expert review during construction and automated checks verify consistency between evidence, execution rules, and scores. Offline programs reconstruct rules and states from public text in the frozen test set; for visual instances, the checks also compare actual attachments and historical registration chains with reviewed image correspondences. We then compare the reconstructed states with private initial states and scoring criteria. All 2240 instances pass these checks, and all reference executions earn full scores. To test scoring beyond the reference paths, prespecified alternative trajectories cover early termination, repeated actions, outdated evidence, incorrect operations, and unnecessary recomputation. All instances also pass checks for consistency between clean and background histories and isolation of specified fields, including answers, optimal trajectories, and private labels. Appendix~\ref{app:scc} and~Appendix~\ref{app:protocol-details} give the scope, examples, and criteria.

\section{Evaluation Protocol and Experimental Setup}
\label{sec:evaluation}

We compare six configurations in a common space-task environment, controlling session, history, and context-window conditions through paired ablations.

\subsection{Agent Framework and Systems Under Evaluation}
We use Pi Agent (Pi), an open-source, lightweight execution framework with potential to be adapted into an embodied spacecraft agent. Pi provides reasoning and tool use, native context management and compaction, and interfaces for external memory. Its simple agent loop limits interference from other design choices, making it a suitable baseline for isolating the effects of memory augmentation. The four augmented configurations share Pi's execution interface and main reasoning model; Appendix~\ref{app:implementation} provides source code and fixed versions.

We use Zhipu's GLM-5.3-Flash for main reasoning in all groups to hold the model constant and focus comparisons on memory systems.

The evaluated systems process history differently. Pi+RAG stores and retrieves historical fragments as a direct-retrieval baseline. Mem0 extracts and integrates important information into persistent memory and updates it~\cite{chhikara2025mem0}. CoM indexes raw interaction nodes and assembles evidence chains after retrieval, linking dispersed instructions, equipment states, and events~\cite{xu2026com}. LightMem organizes history through compression, staged processing, and offline updating~\cite{fang2026lightmem}.

SpaceMind serves as a full-architecture reference for autonomous on-orbit servicing, retaining native recent-observation management, skill routing, and test-time learning~\cite{wu2026spacemind}. Asterisks distinguish its results from Pi-integrated memory modules. Table~\ref{tab:experiment-config} lists shared model, hardware, and runtime settings; Appendix~\ref{app:implementation} gives fixed system parameters and adaptations.
\begin{table*}[tbp]\centering
\caption{Experimental configuration}
\label{tab:experiment-config}
\setstretch{1}
\centering
\newlength{\configItemWidth}
\ifdim\linewidth<\textwidth
  \small\setlength{\tabcolsep}{1.1pt}
  \setlength{\configItemWidth}{34mm}
\else
  \small\setlength{\tabcolsep}{2pt}
  \setlength{\configItemWidth}{48mm}
\fi
\renewcommand{\arraystretch}{1.12}
\begin{tabularx}{\linewidth}{@{}C{\configItemWidth}Y@{}}\toprule
\textbf{Item} & \textbf{Configuration}\\\midrule
Hardware platform & NVIDIA Thor (ARM64)\\
Runtime environment & Docker, Ubuntu 24.04\\
Base runtimes & Python 3.12.3; Node.js 22.23.2\\
Main reasoning model & GLM-5.3-Flash\\
Embedding model & Qwen3-VL-Embedding-2B\\
Embedding dimensions & 2048\\
Embedding input limit & 32,768 tokens\\
Sampling temperature & 0\\
Reasoning settings & Enabled; low effort\\
Context window & 131,072 or 262,144 tokens\\
Historical turns & 30 turns\\
History length with background & Approximately 70K tokens\\
Instances per condition & 2240\\
\bottomrule\end{tabularx}
\end{table*}

\subsection{History Input and Session Protocol}
For each test, we initialize an independent workspace and memory store. After ingesting 30 read-only historical turns sequentially, the agent executes the current task through environment tools, using actual feedback to guide subsequent actions. Scoring uses the first scored terminal state. Between instances, we clear workspaces, skills, and storage to keep histories and memories independent. If infrastructure interrupts a test, the interruption is recorded separately and execution resumes using the same evaluation unit.

Same-session evaluation (\modeS) runs the current task in the history-ingestion session. New-session evaluation (\modeN) starts through the native interface, retaining the instance's persistent memory. The former permits continued use of raw interaction context; the latter tests whether established memory supports task continuation.

Clean and background histories share decisive evidence, initial states, and scoring criteria; routine operational records extend the latter to approximately 70K tokens. Actual requests also include system prompts, tool definitions, assistant responses, and subsequent interactions. To compare how Pi manages this history, we use 128K and 262K windows: 128K triggers native compaction during ingestion, while 262K retains the complete history at this stage. Appendix~\ref{app:protocol-details} and~Appendix~\ref{app:implementation} specify windows, reserved budgets, and compaction parameters.

\subsection{Main Experiments and Paired Ablations}
Main experiments compare task performance and model usage across six configurations with long background histories. Same-session evaluation uses a 128K window; new-session ingestion and execution both use 262K. Within each session mode, we evaluate all six configurations on the same 2240 instances using the same interaction histories and context-window size. Across both modes, this yields 26,880 valid scores. To examine the factors underlying these comparisons, nine ablation conditions use the same test set (Table~\ref{tab:ablation-design}), yielding 20,160 scores. Main and ablation comparisons are paired instance by instance within their respective frozen batches.

Session-boundary comparisons fix clean history and a 262K window. Crossing Pi, RAG, and Mem0 with both session modes gives pairs A/B, H/I, and F/G, holding history and window constant. C/D adds a Pi session comparison with long background histories, also at 262K.

A/C and B/D test routine background at fixed windows and session modes. C/E compares same-session history management with background: complete history at 262K versus native compaction at 128K. Table~\ref{tab:ablation-results} in Section~\ref{sec:results} lists all 14 prespecified comparisons.
\begin{table*}[tbp]
\caption{Ablation conditions. Each group contains 2240 instances.}
\label{tab:ablation-design}
\setstretch{1}
\centering\fontsize{10}{12}\selectfont
\setlength{\tabcolsep}{1pt}\renewcommand{\arraystretch}{1.12}
\begin{tabular*}{\textwidth}{@{\extracolsep{\fill}}ccccc@{}}\toprule
\textbf{Group} & \textbf{Configuration} & \textbf{History} & \textbf{Window} & \textbf{Session}\\\midrule
A & Pi & Clean & 262K & Same-session\\
B & Pi & Clean & 262K & New-session\\
C & Pi & With background & 262K & Same-session\\
D & Pi & With background & 262K & New-session\\
E & Pi & With background & 128K & Same-session\\
\midrule
F & Pi+Mem0 & Clean & 262K & Same-session\\
G & Pi+Mem0 & Clean & 262K & New-session\\
\midrule
H & Pi+RAG & Clean & 262K & Same-session\\
I & Pi+RAG & Clean & 262K & New-session\\
\bottomrule\end{tabular*}
\par\vspace{2pt}\footnotesize Histories with background contain approximately 70K tokens; 128K and 262K denote 131,072 and 262,144 tokens, respectively.
\end{table*}

\subsection{Evaluation Metrics and Statistical Analysis}
Mean score measures task effectiveness; full-score rate measures complete fulfillment. The underlying deterministic scores range from 0--100 and incorporate completion, state, safety, and resource requirements, while full-score rates use the scoring functions' complete-fulfillment flags. We compare systems through mean score differences on matched instances, holding batch, model, context window, history, and session constant.

\begin{samepage}
We define \emph{headroom-normalized gain} as the fraction of the baseline's remaining score gap closed by memory augmentation:
\begin{equation}
G=\frac{M-B}{100-B}\times100\%,\qquad B<100,
\label{eq:gain}
\end{equation}
\end{samepage}
where $M$ and $B$ denote the mean scores of the augmented configuration and native Pi, respectively. $G=50\%$ closes half the remaining gap, $G=0\%$ leaves the mean unchanged, and $G=100\%$ reaches full marks. Reporting this metric alongside scores helps interpret gains across baseline levels. An asterisk after SpaceMind's name marks its full-architecture comparison with Pi.

Model usage $T$ sums input and output tokens across end-to-end generation requests for history processing, current execution, and final maintenance. It includes main reasoning, summarization, memory-assisted generation, and skill routing. We count all input tokens in each request and use server-reported output token counts, without cache discounts. Embeddings are recorded separately. Mean score $S$ and usage define score efficiency $E=10^9S/T$, in points per billion tokens. Simulated task energy is incorporated directly into scores.

To relate aggregate performance to specific memory requirements, we stratify results by 16 subchallenges, nine scenario tasks, and difficulty. Execution trajectories then allow us to examine how historical evidence, selected actions, and scoring consequences relate in individual cases.

We compute 95\% bootstrap confidence intervals (CIs) for mean scores, paired differences, and $G$ by synchronously resampling construction groups within each template. These groups comprise related instances identified during task construction. The intervals describe task-resampling uncertainty under the fixed templates and observed runs. Instances are equally weighted, with 16 per template; resampling preserves equal template weights and condition pairing. Appendix~\ref{app:protocol-details} gives group definitions, statistical parameters, usage accounting, and the random-guessing reference.

\section{Experimental Results and Analysis}
\label{sec:results}

\subsection{Ablation Analysis}
\label{sec:controlled-results}
Table~\ref{tab:ablation-results} groups the 14 prespecified paired comparisons under nine ablation conditions into six themes. Table~\ref{tab:ablation-design} defines conditions; Appendix~\ref{app:detailed-results} gives absolute scores and complete breakdowns.

\begin{table*}[tp]
\caption{The 14 prespecified paired ablation comparisons, each using the same 2240 instances. Values are mean score differences with 95\% intervals.}
\label{tab:ablation-results}
\setstretch{1}
\centering\small
\setlength{\tabcolsep}{1.5pt}\renewcommand{\arraystretch}{1.20}
\begin{tabularx}{\textwidth}{@{}C{22mm}Y@{}}
\toprule[0.9pt]
{\begin{tabularx}{\linewidth}[b]{@{}Y@{}}
\textbf{Comparison theme}\\
\end{tabularx}} & {\begin{tabularx}{\linewidth}[b]{@{}Y C{22.5mm}C{36mm}@{}}
\textbf{Comparison} & \textbf{Contrast} & \shortstack{\textbf{Mean score difference}\\\textbf{[95\% CI]}}\\
\end{tabularx}}\\
\midrule[0.7pt]
{\begin{tabularx}{\linewidth}[b]{@{}Y@{}}
Role of raw history\\
\end{tabularx}} & {\begin{tabularx}{\linewidth}[b]{@{}Y C{22.5mm}C{36mm}@{}}
Clean history: same-session minus new-session & $A-B$ & $+62.02$ $[+60.45,+63.58]$\\
With background: same-session minus new-session & $C-D$ & $+53.88$ $[+52.13,+55.55]$\\
\end{tabularx}}\\
\midrule[0.25pt]
{\begin{tabularx}{\linewidth}[b]{@{}Y@{}}
Benefits of memory integration\\
\end{tabularx}} & {\begin{tabularx}{\linewidth}[b]{@{}Y C{22.5mm}C{36mm}@{}}
Mem0 integration in the same session & $F-A$ & $+4.43$ $[+3.02,+5.89]$\\
RAG integration in the same session & $H-A$ & $+5.50$ $[+4.05,+6.92]$\\
Mem0 integration in a new session & $G-B$ & $+54.61$ $[+52.85,+56.33]$\\
RAG integration in a new session & $I-B$ & $+38.33$ $[+36.49,+40.14]$\\
\end{tabularx}}\\
\midrule[0.25pt]
{\begin{tabularx}{\linewidth}[b]{@{}Y@{}}
Preservation of cross-session execution\\
\end{tabularx}} & {\begin{tabularx}{\linewidth}[b]{@{}Y C{22.5mm}C{36mm}@{}}
Mem0: same-session minus new-session & $F-G$ & $+11.85$ $[+10.19,+13.50]$\\
RAG: same-session minus new-session & $H-I$ & $+29.20$ $[+27.38,+30.98]$\\
\end{tabularx}}\\
\midrule[0.25pt]
{\begin{tabularx}{\linewidth}[b]{@{}Y@{}}
System differences across session boundaries\\
\end{tabularx}} & {\begin{tabularx}{\linewidth}[b]{@{}Y C{22.5mm}C{36mm}@{}}
Same-session: Mem0 minus RAG & $F-H$ & $-1.07$ $[-2.44,+0.34]$\\
New-session: Mem0 minus RAG & $G-I$ & $+16.28$ $[+14.32,+18.26]$\\
\end{tabularx}}\\
\midrule[0.25pt]
{\begin{tabularx}{\linewidth}[b]{@{}Y@{}}
Effects of routine background\\
\end{tabularx}} & {\begin{tabularx}{\linewidth}[b]{@{}Y C{22.5mm}C{36mm}@{}}
Same-session: clean minus background & $A-C$ & $+8.97$ $[+7.39,+10.60]$\\
New-session: clean minus background & $B-D$ & $+0.82$ $[-0.17,+1.84]$\\
Between-session difference in background loss & \shortstack{$(A-C)$\\$-(B-D)$} & $+8.14$ $[+6.27,+10.03]$\\
\end{tabularx}}\\
\midrule[0.25pt]
{\begin{tabularx}{\linewidth}[b]{@{}Y@{}}
Changes in native history management\\
\end{tabularx}} & {\begin{tabularx}{\linewidth}[b]{@{}Y C{22.5mm}C{36mm}@{}}
128K native compaction minus 262K & $E-C$ & $+3.54$ $[+1.75,+5.32]$\\
\end{tabularx}}\\
\bottomrule[0.9pt]
\end{tabularx}
\par\vspace{3pt}
\noindent\begin{minipage}{\textwidth}\small
Table~\ref{tab:ablation-design} defines A--I. Contrasts subtract group means; $E-C$ compares 128K native compaction with 262K complete history. Table~\ref{tab:ablation-details} gives absolute scores.
\end{minipage}
\end{table*}

\paragraph{The role of raw history}
At the same 262K window, Pi scores 62.02 points higher in the same session than after a restart; with routine background, the gap remains 53.88. This substantial history effect shows that the benchmark tests dependence on information retained from prior interactions.

\paragraph{Benefits of integrating memory}
With clean history and a 262K window, Mem0 and RAG improve same-session scores over Pi by 4.43 and 5.50 points, respectively; after restart, these gains reach 54.61 and 38.33. External memory can therefore improve information use even when the complete history remains visible, with greater benefits for continuity once the original context is unavailable.

\paragraph{Preservation of cross-session execution}
In Table~\ref{tab:session-continuity}, Mem0 and RAG reduce Pi's session score gap from 62.02 to 11.85 and 29.20 points, respectively. Both improve continuity, with Mem0 preserving more same-session performance. Persistent memory organization thus affects execution after restart.

\begin{table*}[tbp]
\caption{Session comparison with clean history and a fixed 262K window: 2240 instances per condition. Values are means and same-session minus new-session differences.}
\label{tab:session-continuity}
\setstretch{1}
\centering\small
\setlength{\tabcolsep}{3pt}\renewcommand{\arraystretch}{1.15}
\begin{tabularx}{\textwidth}{@{}Y C{15mm}C{15mm}C{16mm}@{}}\toprule
\textbf{Configuration (group)} & \textbf{Same-session} & \textbf{New-session} & \textbf{Session gap}\\\midrule
Pi (A/B) & 83.12 & 21.10 & 62.02\\
Pi+RAG (H/I) & 88.63 & 59.43 & 29.20\\
Pi+Mem0 (F/G) & 87.56 & 75.71 & 11.85\\
\bottomrule\end{tabularx}
\par\vspace{3pt}\noindent\begin{minipage}{\textwidth}\footnotesize
Differences are calculated from unrounded means; complete paired intervals are provided in Table~\ref{tab:ablation-results}.
\end{minipage}
\end{table*}

\paragraph{System differences across session boundaries}
Mem0 scores 1.07 points below RAG in the same session (95\% interval [$-2.44$,$+0.34$]), consistent with similar performance. After restart, Mem0 leads by 16.28 points [14.32,18.26]. This widening gap shows why same-session performance and cross-session memory require separate evaluation.

\paragraph{Effects of routine background}
Routine operational records lower Pi's same-session score by 8.97 points. In a new session, clean and background histories differ by 0.82 points (95\% interval [$-0.17$,1.84]). Background-related losses differ by 8.14 points across session modes. With decisive evidence unchanged, routine background primarily interferes with same-session execution that still reads raw history, showing the importance of filtering long histories.

\paragraph{Changes in native history management}
\label{sec:history-management}
With background in the same session, Pi scores 3.54 points higher with native compaction at 128K than with complete history at 262K. This aggregate gain masks opposing task effects: downlink confirmation C3 and result reuse R3 improve by 28.65 and 8.88 points, whereas multi-operation state I2 and energy reservation R2 decline by 9.02 and 6.22. These gains and losses show why history management must account for each task's information needs.


\subsection{Main Experiment Analysis}
\label{sec:task-differences}
\begin{figure*}[tp]
\centering\includegraphics[width=.79\textwidth]{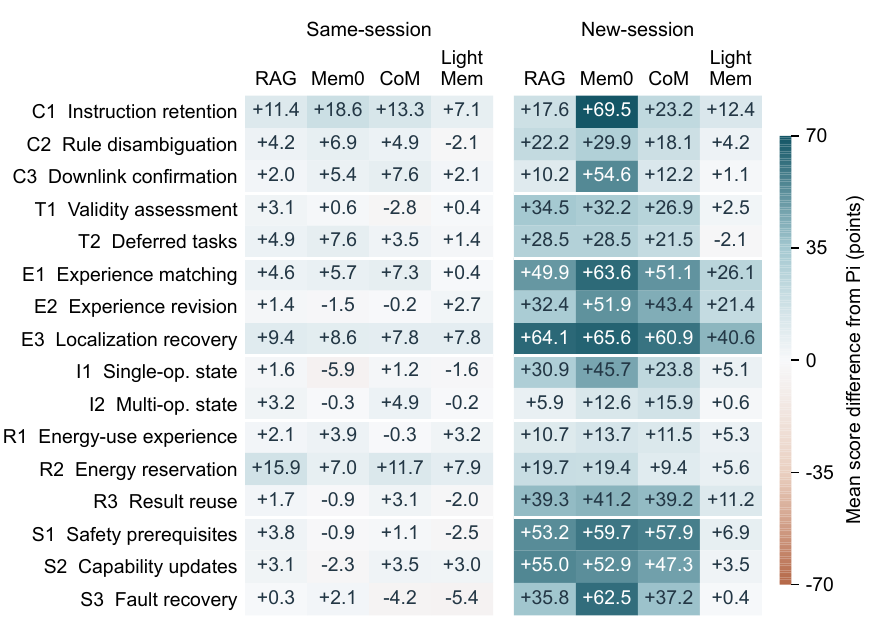}
\caption{Subchallenge score differences from Pi for four memory augmentations under each session condition; positive values indicate gains. RAG, Mem0, CoM, and LightMem augment the same Pi framework. Conditions follow Table~\ref{tab:main-results}. Each column has 2240 instances: 128 each for E3 and I1, 112 for I2, and 144 for each remaining subchallenge. Table~\ref{tab:main-domain-results} gives absolute scores.}
\label{fig:main-memory-profile}
\end{figure*}

Main experiments compare six configurations with long background histories (Table~\ref{tab:main-results}). Figure~\ref{fig:main-memory-profile} shows Pi-relative score changes across 16 subchallenges for the four Pi-integrated memory systems.

\begin{table*}[tp]
\caption{Main results for six configurations: 128K same-session and 262K new-session evaluation, with 2240 instances per row and 26,880 scores in total. Bold marks session-wise best values excluding Pi and SpaceMind: lowest tokens and highest mean score, full-score rate, $G$, and efficiency.}
\label{tab:main-results}
\setstretch{1}
\centering\small
\setlength{\tabcolsep}{2pt}\renewcommand{\arraystretch}{1.22}
\setlength{\aboverulesep}{2pt}\setlength{\belowrulesep}{2pt}
\begin{tabularx}{\textwidth}{@{}C{25mm}C{39mm}Y Y Y Y@{}}
\toprule[1.1pt]
\textbf{Configuration} & \textbf{Mean score}\newline\textbf{[95\% CI]} & \shortstack{\textbf{Full-score}\\\textbf{rate (\%)}} & \textbf{$G$/\%} & \textbf{Tokens}\newline\textbf{(M)} & \textbf{Score}\newline\textbf{efficiency} $E$\\
\midrule[0.8pt]
\multicolumn{6}{c}{\textbf{Same-session (128K)}}\\

Pi & 78.20 [76.80,79.57] & 69.82 & 0.00 & 2721.58 & 28.73\\
Pi+\textbf{RAG} & \textbf{82.74} [81.45,83.99] & \textbf{75.71} & \textbf{20.82} & 3089.99 & \textbf{26.78}\\
Pi+\textbf{Mem0} & 81.70 [80.38,82.99] & 74.60 & 16.07 & 3982.69 & 20.51\\
Pi+\textbf{CoM} & 82.07 [80.79,83.34] & 74.64 & 17.75 & 3585.64 & 22.89\\
Pi+\textbf{LightMem} & 79.59 [78.17,80.96] & 71.38 & 6.38 & \textbf{3077.36} & 25.86\\
SpaceMind$^{*}$ & 44.10 [42.50,45.68] & 32.77 & -156.41 & 1945.16 & 22.67\\
\midrule
\multicolumn{6}{c}{\textbf{New-session (262K)}}\\

Pi & 21.03 [19.85,22.26] & 11.74 & 0.00 & 2744.22 & 7.66\\
Pi+\textbf{RAG} & 53.03 [51.47,54.59] & 43.48 & 40.53 & 3359.77 & 15.78\\
Pi+\textbf{Mem0} & \textbf{65.28} [63.66,66.88] & \textbf{55.45} & \textbf{56.03} & 4119.77 & \textbf{15.85}\\
Pi+\textbf{CoM} & 52.32 [50.68,53.97] & 42.99 & 39.62 & 3926.32 & 13.32\\
Pi+\textbf{LightMem} & 30.00 [28.57,31.37] & 20.58 & 11.36 & \textbf{3156.11} & 9.51\\
SpaceMind$^{*}$ & 20.49 [19.36,21.60] & 10.71 & -0.68 & 1842.54 & 11.12\\
\bottomrule[1.1pt]
\end{tabularx}
\par\vspace{3pt}
\noindent\begin{minipage}{\textwidth}\small
Full-score rate measures complete fulfillment. $G=(M-B)/(100-B)\times100\%$, with $B$ the corresponding Pi mean; intervals use synchronized paired resampling of construction groups within templates. $^{*}$ marks SpaceMind's full architecture, with its native budgets, usage control, and scheduling. Total tokens cover completed generation requests with full usage records, including main reasoning, summarization, memory-assisted generation, and skill routing; embeddings are counted separately. Efficiency $E=10^9S/T$ is in points per billion tokens; 1M tokens=$10^6$ tokens.
\end{minipage}
\end{table*}

\paragraph{Same-session performance and cost}
Native Pi scores 78.20 with its own compaction; adding RAG raises the score to 82.74. Mem0, CoM, and LightMem score 81.70, 82.07, and 79.59, respectively: gains of 1.39--3.87 over Pi, but still below RAG. These gains also incur greater cost: relative to Pi, end-to-end token usage rises by 13.1\%--46.3\%, with efficiencies of 20.51--25.86 points per billion tokens, below Pi's 28.73 and RAG's 26.78. Full-architecture reference SpaceMind scores 44.10. Native compaction and direct retrieval thus provide competitive performance and efficiency in the same session, while the benchmark exposes the score--cost tradeoffs introduced by different memory mechanisms.

\paragraph{Cross-session benefits of knowledge organization and maintenance}
After restart, Pi scores 21.03; RAG, Mem0, CoM, and LightMem score 53.03, 65.28, 52.32, and 30.00; SpaceMind scores 20.49. Mem0 leads RAG by 12.25 points (95\% interval [10.09,14.37]). Compared with RAG, it increases the full-score rate from 43.48\% to 55.45\% and achieves the highest new-session score in all nine tasks (Table~\ref{tab:main-scenario-results}). CoM performs similarly to RAG overall but leads on experience revision E2, scoring 69.64 versus 58.55. LightMem scores below RAG overall. Knowledge organization and maintenance can therefore improve on direct retrieval, but the benefits depend on the implementation and task.

\paragraph{Possible fact-centered and retrieval-oriented bias}
Figure~\ref{fig:main-memory-profile} shows clear new-session benefits for instruction retention C1, experience revision E2, and result reuse R3. Across Mem0, CoM, and LightMem, Pi-relative gains range from 12.4--69.5 points for C1, 21.4--51.9 for E2, and 11.2--41.2 for R3. Experience matching E1 improves by 26.1--63.6 points; deferred tasks T2 improve by 28.5 with Mem0 and 21.5 with CoM. Localization recovery E3 and safety prerequisites also benefit from recovering key historical evidence. Correctly preserving and recalling valid instructions, change records, existing results, and deferred tasks can directly constrain actions.

In contrast, ongoing state maintenance and resource planning remain weaker. RAG, Mem0, and CoM exceed 73 points on result reuse R3 but reach at most 49.94 on multi-operation state I2 and 40.63 on energy reservation R2. Remembering existing values or future tasks differs from continuously tracking completed steps, unfinished work, and resource commitments. This distinction also helps interpret the E2 gains, which concern judgment revision from recorded changes. Table~\ref{tab:case-studies} links historical evidence, actual actions, and resource consequences.

We therefore hypothesize a fact-centered, retrieval-oriented inductive bias: the tested systems favor structured retention, updating, and recall of salient facts, explicit constraints, and key contextual anchors, with less sustained support for dynamic states, procedural learning from experience, planning, and closed-loop execution.

\paragraph{Anomaly analysis}
SpaceMind scores 44.10 in the same session, below Pi's 78.20. Its native configuration suggests a mismatch between task demands and execution mechanisms. Designed for local perception--action loops in autonomous on-orbit servicing~\cite{wu2026spacemind}, it allows at most three ReAct rounds per internal decision step and one tool call per round. Main reasoning is limited to 1024 generated tokens per call, versus Pi's 16,384. Its history manager keeps detailed observations from the latest eight steps and truncates and archives earlier ones (Appendix~\ref{app:implementation}). On this benchmark's diverse space tasks, these specialized budgets and retention policies may limit long-history evidence linking and sustained execution, helping explain the low same-session score.

We retain SpaceMind's native round budgets, history management, and scheduling to evaluate its official architecture as configured. Its results provide a system-level reference for assessing how an on-orbit servicing architecture transfers to the benchmark's broader space-task requirements.

\subsection{Design Implications for Spacecraft Memory Systems}
\label{sec:design-implications}

The subchallenge results and execution cases suggest four directions for memory design in embodied spacecraft agents.

\paragraph{Valid states under interruption, interference, and updates}
Results for instruction retention, rule disambiguation, and multi-operation state suggest that preserving requirements and maintaining complex states need distinct designs. As telemetry changes and multistep operations alter physical states, memory must track valid instructions, progress, and unfinished work even when ground instructions are interrupted or arrive out of order. Instruction validity depends on issuance and update relationships, while operational states must be revised from actual feedback, with evidence retained to distinguish plans, attempts, and completion. Maintaining these distinctions supports resumption at the correct point after interruption and timely judgment revision as updates arrive.

\paragraph{Supporting actions proactively through task patterns}
In the repair case, RAG retrieves relevant records but uses another device's test-run feedback to judge the current device's repair, showing the need for object matching and applicability checks after retrieval (Table~\ref{tab:case-studies}). With limited onboard computation, memory systems could organize instructions, states, safety prerequisites, and experience around inspection, repair, and replenishment patterns, supplying applicable memory before action selection. Stable procedures for recurring information filtering, evidence linking, and prerequisite checking may ease the reasoning burden of detailed tasks and guide current reasoning and action.

\paragraph{Learning transferable experience with explicit applicability conditions}
Gains in experience matching and revision show the value of prior experience and records of subsequent changes. Mem0 organizes this information through extraction and updating, whereas CoM retains raw records and assembles evidence chains~\cite{chhikara2025mem0,xu2026com}. Over long-term operation, memory systems could distill task methods, conditional judgments, and revision principles, including which prerequisites need reconfirmation and which feedback calls for strategy adjustment. Such higher-level experience could support transfer across tasks, whereas specific parameters and computed results must remain linked to their equipment, configuration, and operating conditions. Reassessing these conditions before reuse would prevent incidental values from one successful task from becoming general rules. Experience summaries should preserve applicability conditions and the evidence behind revisions, allowing later feedback to update the stored experience.

\paragraph{Organizing memory around resource budgets and time windows}
The new-session performance gap between result reuse and energy reservation suggests different memory organization needs. Beyond grouping records by object, type, and time, memory could link progress, later steps, deadlines, expected consumption, and reserved resources to task goals and execution conditions. This would clarify which results can be reused, which resources are committed to later work, and whether current choices affect completion within future windows. Memory organization itself must also fit computation and time budgets, with update scope and maintenance timing matched to task urgency.

\section{Conclusion}

AstroMemBench provides domain-specific memory evaluation for embodied spacecraft agents. Its three-dimensional Scenario--Challenge--Capability method organizes operational requirements into 140 templates and 2240 executable instances, linking historical information to task completion, state changes, safety, and resource use through deterministic scoring.

Nine ablation conditions and 14 paired comparisons validate history dependencies and the benchmark's discriminative ability. Main experiments show that configurations with similar same-session performance differ in new-session evaluation. The tested systems support instruction retention and result reuse more strongly than ongoing state tracking and energy reservation.

With common tasks and scoring rules, these evaluations identify distinct needs for historical retrieval, state maintenance, and resource planning, providing a basis for memory system design, improvement, and selection.

AstroMemBench 1.0 is publicly available on GitHub. Future versions will refine the benchmark for long-term continuous tasks and richer multimodal interactions.

\FloatBarrier
\section*{Author Contributions}
\textbf{You~Wu:} Conceptualization, Methodology, Software, Validation, Formal analysis, Investigation, Data curation, Writing -- original draft, Writing -- review \& editing, Visualization.
\textbf{Guhong~Zhang:} Conceptualization, Methodology, Validation, Resources, Visualization, Supervision.
\textbf{Hongwei~Li:} Methodology, Validation, Writing -- review \& editing, Supervision.
\textbf{Xinyue~Hong:} Validation, Writing -- review \& editing, Visualization.
\textbf{Yuan~Jia:} Resources, Supervision, Project administration.

\section*{Funding}
The authors received no specific funding for this research.

\section*{Conflict of Interest}
The authors declare no competing financial interests or personal relationships that could have influenced the work reported in this paper.

\section*{Data availability}
AstroMemBench 1.0 and its evaluation code are publicly available at \url{https://github.com/flyingView/AstroMemBench-v1}.

\section*{AI Assistance Disclosure}
OpenAI Codex was used for translation, language editing, and manuscript and figure formatting. The authors take full responsibility for the content of this manuscript.

\label{main:end}
\bibliographystyle{elsarticle-num}
\begingroup\raggedright
\bibliography{references}
\endgroup
\clearpage
\appendix
\numberwithin{table}{section}
\numberwithin{figure}{section}
\section{Criteria and Evidence for Benchmark Comparison}
\label{app:related-work}

Table~\ref{tab:benchmarks} compares 13 memory benchmarks and includes AstroAgentBench as a space-domain reference. This appendix defines the comparison criteria and examines each benchmark's protocols and coverage evidence to establish when and how history affects the current task.

\subsection{Criteria for Capability Coverage}
\label{app:coverage-criteria}
Coverage judgments are based on tasks, procedures, and metrics reported in the papers. Operational definitions of the five general capabilities and six space-relevant capabilities are provided in Table~\ref{tab:coverage-definitions}. \yes{} indicates direct measurement, \partly{} a related measurement, and \none{} no corresponding measurement identified. A representative task suffices for coverage of a column, and columns are assessed independently. Conference versions are preferred. PM-Bench uses the accepted author manuscript, while AstroAgentBench uses the v3 author manuscript marked AACL-IJCNLP 2026 and is marked with an asterisk to identify it as a spacecraft-agent benchmark.

{
\setstretch{1}
\fontsize{10}{12}\selectfont
\begin{table*}[tp]
\centering
\caption{Operational definitions of comparison columns and their correspondence to AstroMemBench subchallenges}\label{tab:coverage-definitions}
\begin{tabularx}{\textwidth}{C{22mm}Y C{25mm}}
\toprule[0.9pt]
Comparison column & Criteria for direct and partial coverage & Corresponding subchallenges\\\midrule[0.7pt]

Factual recall & Historical objects, attributes, events, or parameters support current answers or actions. & All six categories\\
Temporal association & Uses event order, counts, or duration; merely retaining history in order counts as partial coverage. & C2, T, I2\\
State maintenance & Accumulates task states, uses updated facts, or assesses the applicability of prior experience. & C3, E2, I1/I2\\
Visual memory & Uses objects, attributes, or locations in earlier images to determine current judgments. & E3-P1/P2\\
Experience transfer & Applies rules or skills to new objects, starting points, or subsequent tasks; matching familiar cases, reusing preferences, and exact imitation count as partial coverage. & E1 (partial)\\
Instruction validity & Uses history to retain or identify currently valid instructions; inferring persistent rules only from demonstrations counts as partial coverage. & C1/C2, T1\\
Action windows & A historical time interval or event condition determines when to act; scheduling dates in a plan alone counts as partial coverage. & T1/T2\\
Unknown environments & Uses historical observations or probing to act under unknown or unobserved states; incremental information acquisition through textual tools counts as partial coverage. & E1--E3\\
Physical interaction & Prior physical effects or interaction feedback determine subsequent operations; navigation alone counts as partial coverage. & I1/I2\\
Resource cost & Energy or computation cost within the task affects choices or scores; path or monetary costs count as partial coverage. & R1--R3\\
Safety and reliability & Historical safety prerequisites, interaction failures, or recovery progress determine subsequent reliable operations. & S1--S3\\
\bottomrule
\end{tabularx}
\end{table*}

}

\subsection{History Reuse, Environment Interaction, and Coverage Evidence}
Tables~\ref{tab:benchmark-protocols} and~\ref{tab:benchmark-protocols-action} summarize general history-understanding, task-execution, and embodied protocols. Coverage judgments use representative tasks to distinguish experience collection, offline answering, and execution in the current environment.

{
\setstretch{1}
\fontsize{10}{12}\selectfont
\setlength{\tabcolsep}{3pt}\renewcommand{\arraystretch}{1.00}
\begin{table*}[tp]
\fontsize{10}{11.2}\selectfont
\captionsetup{skip=4pt}
\centering
\caption{Protocols and key coverage evidence for general history-understanding benchmarks. Source locations refer to sections, figures, tables, or appendices in the original papers.}\label{tab:benchmark-protocols}
\begin{tabularx}{\textwidth}{@{}C{35mm}C{38mm}C{30mm}Y@{}}
\toprule[0.9pt]
\textbf{Benchmark and source location} & \textbf{History and current task} & \textbf{Reuse and interaction} & \textbf{Evidence for capability coverage}\\
\midrule[0.7pt]

LoCoMo~\cite{maharana2024locomo}\newline\S4.1/4.3; Fig.~2 & Multisession text and images; QA, summarization, and generation & Predetermined session histories; offline outputs & Retrospective temporal QA, state generation, and historical photographs support the corresponding memory capabilities; preference reuse is related to transfer.\\
LongMemEval~\cite{wu2025longmemeval}\newline\S3; Fig.~1; Table~1 & Multisession text; QA and abstention & Predetermined session histories; offline answers & Extraction, temporal reasoning, and knowledge updating directly test the first three general capabilities; recommendations based on historical preferences are related to transfer.\\
MemoryAgentBench~\cite{hu2026memoryagentbench}\newline\S3.1--3.3 & Incremental materials; QA, classification, and summarization & Incremental reading of the same materials; offline outputs & EventQA tests order, FactConsolidation tests factual updates, and test-time classification tests transfer.\\
Mem-Gallery~\cite{memgallery}\newline\S3.3; Appendix~A.3; Fig.~16 & Multisession text and images; QA and evidence retrieval & Predetermined session histories; offline outputs & Conflict resolution and visual reasoning test state and visual memory; classification of new images tests transfer; efficiency is measured as system runtime.\\
LoCoMo-Plus~\cite{li2026locomoplus}\newline\S3--4; Appendix~C & Conversations with implicit constraints; constraint-consistent answers & Predetermined session histories; offline answers & The complete framework retains factual and temporal tasks; persistent goals and emotions test state; adherence to implicit constraints is related to instruction retention.\\
AMA-Bench~\cite{amabench}\newline\S3.1--3.3; Figs.~4--5 & Real and synthetic trajectories; experience QA & A single existing trajectory; offline answers & Trajectory QA tests recall, time, and updates through questions about historical actions.\\
\bottomrule
\end{tabularx}
\end{table*}

}

{
\setstretch{1}
\fontsize{10}{12}\selectfont
\setlength{\tabcolsep}{3pt}\renewcommand{\arraystretch}{1.00}
\begin{table*}[tp]
\fontsize{10}{11.2}\selectfont
\captionsetup{skip=4pt}
\centering
\caption{Protocols and key coverage evidence for task-execution and embodied benchmarks and the space-domain reference.}\label{tab:benchmark-protocols-action}
\begin{tabularx}{\textwidth}{@{}C{35mm}C{38mm}C{30mm}Y@{}}
\toprule[0.9pt]
\textbf{Benchmark and source location} & \textbf{History and current task} & \textbf{Reuse and interaction} & \textbf{Evidence for capability coverage}\\
\midrule[0.7pt]

Mem2ActBench~\cite{shen2026mem2actbench}\newline\S3.3--4.1; Tables~8--10 & Multisession text; tool and argument generation & Predetermined histories; offline call generation & Evolution chains test updated arguments and requirements; dates and budgets are parameters recovered for tool calls.\\
PM-Bench~\cite{pmbench}\newline\S3.1--3.3 & Schedules, events, and queries; actions when due & The same situation across days; queries and execution & Cancellation and rescheduling test valid intentions; time windows and event triggers test action timing; queries through hidden channels are related to information acquisition in unknown environments.\\
MemoryArena~\cite{he2026memoryarena}\newline\S3.1; Fig.~2 & Own execution and feedback; task completion & Interdependent subtasks; tool interaction & Prior requirements constrain later choices; travel dates, tool searches, and monetary budgets are related to action windows, unknown environments, and resource cost, respectively.\\
FindingDory~\cite{findingdory}\newline\S3; Table~2 & Images, poses, and actions; navigation & Navigation after prescribed experiences & Historical attributes, rearrangements, and images test four general capabilities; unobservable locations support unknown-environment coverage; navigation is related to physical interaction.\\
LMEE-Bench~\cite{wang2026lmee}\newline\S3--5; Figs.~3--4 & Exploration images and locations; navigation and QA & Exploration and execution in the same scenario & Visit progress and images support state and visual memory; SPL is related to resource cost; instructions are issued incrementally during exploration.\\
MIKASA-Robo~\cite{mikasa}\newline\S6; Appendix~K & Images and robot states; physical operations & A single interaction process; current operations & Color sequences, relative orientations, and stage progress test memory; exact imitation is related to transfer; precision constraints govern current operations.\\
RoboMME~\cite{dai2026robomme}\newline\S3; Tables~1--2; Appendix~E & Images, demonstrations, and histories; physical operations & History and demonstrations within a single task; current operations & Counting, occlusion, and demonstrations test general memory; StopCube tests brief windows; reported resource metrics include system tokens and TFLOPs.\\
AstroAgentBench$^{*}$~\cite{wang2026astroagentbench}\newline\S3, 5.3; Appendices~C, I.2 & Training notes and validation feedback; space planning & Notes frozen after training; code and plan validation & Tests experience transfer and energy constraints; planning windows and validation feedback are related to the corresponding capabilities; onboard actions are reconstructed in the planning model.\\
\midrule[0.25pt]
\textbf{AstroMemBench} & \textbf{Task records and images; space operations} & \textbf{Isolated histories per instance; current actions and state transitions} & \textbf{Uses common histories and environments to examine task completion, state, safety, and resource consequences; distinguishes same-session and new-session continuity.}\\
\bottomrule
\end{tabularx}
\end{table*}
\par\vspace{3pt}\footnotesize
$^{*}$ denotes a spacecraft-agent benchmark. SPL is the success rate weighted by path length. System processing overhead and resource costs within tasks are assessed separately.
}

\FloatBarrier
\section{Task Construction and Consistency Validation}
\label{app:scc}

\subsection{Applicability and Capability Coverage}
Before checking task applicability, SCC specifies what history each subchallenge requires, when it is needed, and what happens if it is misused. Four of 144 combinations fall outside the task definitions: E3$\times$D3 would require communication sensors for auxiliary navigation; I1$\times$D3 would require repair through disassembly and reassembly; I2$\times$O1 would require additional material and operating-condition assumptions for accumulated damage; I2$\times$D3 would treat byte accumulation as a cumulative physical effect, although it concerns data processing. Excluding these combinations leaves 140 templates, one per applicable combination.
\label{app:construction-details}

Template identifiers use FM--subchallenge--scenario task, e.g., FM-C1-P1. Capability labels reflect actual perceptual, navigational, operational, and planning behaviors, allowing multiple labels per template. The 140 templates form 285 associations (Fig.~\ref{fig:associations}). Capabilities describe test-set coverage; execution consequences determine scores.

\subsection{From History Dependencies to Execution Criteria}
Each subchallenge explicitly links history to current actions: history establishes current rules and states, applicable experience, or pending tasks, and public observations and action interfaces let the agent use this information. State transitions and scoring then determine whether the agent has used it correctly. Table~\ref{tab:subchallenge-specifications} summarizes criteria for all 16 subchallenges.

Each instance contains history, the initial environment state and public observation, action-dependent state-transition rules, and a scoring function. After ingesting 30 historical turns sequentially through its own memory mechanisms, the agent executes using only public observations and actual feedback; answers, reference actions, scores, and challenge labels remain private.

\begin{table*}[tp]
\caption{History dependencies and execution criteria for 16 subchallenges. Rows give common template requirements; actions, partial scores, and termination are instance-specific.}
\label{tab:subchallenge-specifications}
\centering\normalsize
\setlength{\tabcolsep}{3pt}\renewcommand{\arraystretch}{1.08}
\begin{tabularx}{\textwidth}{@{}C{27mm}Y Y Y@{}}
\toprule[0.9pt]
\textbf{Subchallenge} & \textbf{History to retain} & \textbf{Current use} & \textbf{Execution and scoring criteria}\\
\midrule[0.7pt]
C1 Instruction retention & Still-valid requirements from before an interruption & Continue without new ground confirmation & Meet all applicable requirements and constraints.\\
C2 Rule disambiguation & Instruction issuance, replacement, revocation, and applicability & Resolve conflicting or out-of-order records & Follow valid rules, rather than reception order.\\
C3 Downlink confirmation & Transmission attempts, targets, and reception acknowledgments & Assess delivery completion & Distinguish requests from delivery; correctly resume, retransmit, or terminate.\\
\midrule[0.25pt]
T1 Validity assessment & Activation and expiration conditions for rules, data, or credentials & Assess current applicability & Use information only within valid periods or conditions.\\
T2 Deferred tasks & Accepted tasks and their time or event triggers & Recall tasks when triggered & Execute retained requirements within their timing constraints.\\
\midrule[0.25pt]
E1 Experience matching & Experience under known environments or conditions & Match current objects and conditions & Apply suitable experience without misusing similar records.\\
E2 Experience revision & Prior experience and later applicability changes & Revise judgments using recorded changes & Use updated, rather than invalidated, experience.\\
E3 Localization recovery & Associations among prior visits, images, and locations & Recover location without current location labels & Identify the current location and act accordingly.\\
\midrule[0.25pt]
I1 Single-operation state & Connection, assembly, or other operations and feedback & Reconstruct physical relationships and the next step & Distinguish plans, success, and failure; match actual states.\\
I2 Multi-operation state & Progress and cumulative physical effects of multiple steps & Track states across objects and steps & Respect progress and dependencies; avoid repeated or conflicting operations.\\
\midrule[0.25pt]
R1 Energy-use experience & Energy use and safe traversal under known conditions & Plan transport, operations, or resource use & Jointly assess completion, energy use, and hazardous or blocked actions.\\
R2 Energy reservation & Subsequent tasks and outstanding energy needs & Reserve energy for future tasks & Complete the current task while meeting subsequent electrical-energy commitments.\\
R3 Result reuse & Computed results and their equipment, configurations, and conditions & Check whether results remain reusable & Submit correctly while reducing additional simulated computation energy.\\
\midrule[0.25pt]
S1 Safety prerequisites & Valid safety evidence for the current object & Confirm prerequisites before action & Satisfy safety requirements through both actions and submitted evidence.\\
S2 Capability updates & Functional changes, disabling, and restoration & Use currently valid capabilities & Avoid invalidated capabilities and meet changed constraints.\\
S3 Fault recovery & Progress, repairs, and feedback before and after faults & Resume from the currently valid stage & Correctly assess completion; match actions and evidence to the current object and stage.\\
\bottomrule
\end{tabularx}
\end{table*}

\subsection{Paired Instances: Different Correct Actions Under Identical Current Observations}
Table~\ref{tab:history-pair-example} expands the supply-line example. Both instances share the current request, action menu, and 29 historical turns. Turn 7 records either a successful connection or an unexecuted plan, producing different current states and correct actions. Public feedback does not reveal the true connection state, so the agent must infer it from history before submission. The pair isolates how the critical historical event affects actions and scores.

{
\setstretch{1}
\fontsize{10}{12}\selectfont
\setlength{\tabcolsep}{4pt}\renewcommand{\arraystretch}{1.10}
\begin{table*}[tp]
\centering
\caption{FM-I1-O3-001 and 002: historical evidence determining state. Both histories have 30 turns and differ at turn 7.}\label{tab:history-pair-example}
\begin{tabularx}{\textwidth}{@{}C{27mm}Y Y@{}}
\toprule[0.9pt]
\textbf{Component} & \textbf{FM-I1-O3-001} & \textbf{FM-I1-O3-002}\\
\midrule[0.7pt]

Operation rules & \multicolumn{2}{C{\dimexpr\textwidth-27mm-4\tabcolsep\relax}}{H1 and H2 can each connect to at most one port. To retract a target line, first disconnect it from its actual port if it is still connected; otherwise, stow it directly. Plans do not change physical states, and the other line remains in place.}\\
Shared history & \multicolumn{2}{C{\dimexpr\textwidth-27mm-4\tabcolsep\relax}}{Turn 2: H1 connects to P01 and H2 to P04. Turn 5: H1 is successfully disconnected from P01. Turns 10 and 27: plans to connect H1 to P05 are proposed but not executed. Subsequent H2 operations are identical in both instances, with a final connection to P01.}\\
Difference in turn 7 & Executes connection of H1 to P02; feedback indicates success; H1 is currently connected to P02. & Proposes a plan to connect H1 to P02; no physical operation has begun, and existing connections are unchanged.\\
Current request & \multicolumn{2}{C{\dimexpr\textwidth-27mm-4\tabcolsep\relax}}{Begin retracting H1 and submit the next operation; pump shutdown, isolation, and depressurization are already complete.}\\
Public action menu & \multicolumn{2}{C{\dimexpr\textwidth-27mm-4\tabcolsep\relax}}{Both instances have the same 12-option menu: disconnect H1 or H2 from one of P01--P05, or stow the line. The menu is not filtered by current state. Option 7 disconnects H1 from P02; option 4 stows H1.}\\
State reconstructed from history & H1 is connected to P02 & H1 is not connected to any port\\
Correct submission & \texttt{select}, \texttt{option="7"} & \texttt{select}, \texttt{option="4"}\\
Action and score & Disconnecting H1 from P02 earns 100 points; directly stowing H1 earns 0. & Directly stowing H1 earns 100 points; disconnecting H1 from P02 earns 0.\\
Execution endpoint and feedback & \multicolumn{2}{C{\dimexpr\textwidth-27mm-4\tabcolsep\relax}}{The first valid selection ends execution. At most 3 total submissions are allowed, with format errors consuming the allowance. Both instances return only "The operation request has been recorded," without disclosing the true current connection state.}\\
\bottomrule
\end{tabularx}
\end{table*}

}

\subsection{Background and Visual Materials}
Histories with routine background target 70,000 tokens (range 69,850 to 70,150), including core facts, public wrappers, and tokens accounted for by images. The background draws on 513 reviewed synthetic operational records covering communication, testing, maintenance, and data delivery. For each instance, complete paragraphs are selected in a fixed order and distributed evenly across 30 turns, with paired instances sharing both content and positions. Clean histories remove only this routine background, retaining decisive evidence, task-related distractors, and turn boundaries.

E3 has eight templates and 128 instances. P1 and P2 each have 16 NASA-image instances; the other 96 use textual geometry, distances, clocks, or occlusion. The 13 source files yield 13 main images and 9 P1 references. Configurations receive identical bytes and ordering, without location answers, source filenames, evaluation annotations, OCR, or auxiliary captions. Images retain original dimensions or are proportionally resized: regular main-image long sides are 1536/1280 pixels, two ultrawide panoramas are 4096, and P1 references are 640. Images are neither cropped nor generated. Main experiments and ablations share the frozen images.

\subsection{Evidence Reconstruction and Alternative-Trajectory Validation}
We review the instances and use automated checks to verify consistency among public evidence, environment states, and scoring rules. For 2208 nonvisual instances, programs reconstruct valid rules, message states, physical relationships, safety evidence, and resource constraints from public histories and observations. For 32 visual instances, the checks first validate attachments against reviewed image correspondences, then reconstruct states along visit and location registration chains. All 2240 instances pass consistency checks, and their reference actions earn full scores.

Beyond reference actions, alternative trajectories test equivalent correct expressions, early termination, repeated actions, incorrect evidence, outdated states, unnecessary recomputation, and insufficient resources; all scores follow the specified rules. Reconstructed facts, reference actions, and turn boundaries match between each instance's clean and background histories. Visibility checks across histories, current observations, and feedback also verify isolation of answers, scores, optimal trajectories, challenge and capability labels, and visual source metadata.

\FloatBarrier
\section{Scoring, Statistics, and Runtime Accounting}
\label{app:protocol-details}

\subsection{Scoring Units and Execution Endpoints}
A scoring unit combines one instance, configuration, experimental condition, and repetition, with its own workspace, skills, and memory. Section~\ref{sec:evaluation} describes sessions; Appendix~\ref{app:implementation} specifies context and memory parameters.

Scoring uses the first scored terminal state, with task rules governing format failures, refusals, and missing actions. After an infrastructure interruption, we retain completed units, clear unfinished units' native resources, and resume execution while preserving all attempts and usage records. Requests with unverifiable or exceeded budgets are marked separately as invalid observations.

\subsection{Operational Consequences and Full-Score Conditions}
Scoring maps task consequences to $[0,100]$. For inspection instances, scoring measures the fraction of requirements met; for decision instances, it checks actions and required evidence; for continuous interaction, it checks completion states, errors, and resource costs. Full-score status follows complete-fulfillment conditions in the scoring logic, regardless of display rounding.

In FM-R1-P3, the agent reconstructs safe routes and energy needs from history. Let $E_{\min}$ denote minimum energy for safe transport, $E$ actual energy use, and $b,h$ blocked and hazardous traversal counts. When delivery is complete,
\begin{equation}
s=\max\!\left\{0,100\min\!\left(1,\frac{E_{\min}}{E}\right)-25b-15h\right\}.
\end{equation}
Incomplete delivery earns 0; full marks require safe arrival at minimum energy cost.

FM-R3-O2 scores the fraction of the initial 300 computation-energy units remaining after correct submission. Reusing historical parameters earns 100 points, whereas each requested computation consumes 30 simulated units; correct submissions after one, two, or three requests therefore earn 90, 80, or 70 points. Incorrect or missing submissions earn 0. The scoring thus distinguishes identical correct answers by their resource costs.

\subsection{Equal-Weight Aggregation and Paired Resampling}
Each condition has $N=2240$ instances, 16 per template. Mean score and full-score rate are
\begin{equation}
M=\frac{1}{N}\sum_{i=1}^{N}s_i,\qquad
P_{\mathrm{full}}=\frac{100}{N}\sum_{i=1}^{N}f_i.
\end{equation}
Here $f_i=1$ if instance $i$ fulfills the scoring function's full-score condition, and $f_i=0$ otherwise. Headroom-normalized gain (Eq.~\eqref{eq:gain}) uses Pi's mean under the same condition, retaining negative values when $M<B$ and remaining undefined when $B=100$.

A construction group contains instances explicitly linked during task design, for example by a shared counterfactual setup or image--location mapping; unlinked instances form singleton groups. Groups are sampled with replacement within each template, with all members retained together.

For the 95\% bootstrap intervals, we synchronously resample 2089 construction groups within 140 templates. For $K$ templates, when template $m$ yields $n_m$ sampled instances, each has weight $a_i=1/(Kn_m)$, preserving equal template weights in $\sum_i a_i s_i$. The same construction groups are sampled across conditions; scores, differences, and gain numerators and denominators are recalculated. We use seed 1729, 10,000 resamples, and 2.5\%/97.5\% quantiles. Equal original template sizes make the point estimate identical to the ordinary mean. Each instance has one scored run per condition. The intervals quantify task-resampling uncertainty within the fixed template collection, treating construction groups as exchangeable within each template.

\subsection{Generation Requests and Token Accounting}
Main-experiment usage includes completed generation requests from valid scoring attempts with complete server-reported counts, including refusals and output-limit exhaustion. Input counts cover complete requests; output counts use server reports without recounting reasoning tokens.

End-to-end usage covers preparation, history processing, session switching, execution, and final maintenance, including main reasoning, summarization, memory-assisted generation, and skill routing; embeddings are recorded separately. Per-instance averages are totals divided by 2240; the main table reports group totals and score efficiency.

\subsection{Calculating the Random Full-Score Reference}
\label{app:guessing}
Table~\ref{tab:guessing-counts} uses a specified random strategy without history access. For reporting instances, the strategy samples uniformly from valid nontermination options; for C2, it first performs the unique preparatory operation. For evidence-based instances, it also samples evidence independently and uniformly from all normalized options. Fixed sequences are sampled stepwise; interactive instances follow public menus to a true terminal state, retaining redundant and hazardous action costs. Numerical instances receive a directory of distinct answers from the same template's 16 instances and uniformly select one complete scalar or vector for direct submission. Original rules govern equivalent answers, additional measurements, resource use, and termination.

Each instance's full-score probability $p_i$ comes from exact enumeration or finite-state probability propagation. For 16 independent random answers within a template, the generating function of full-score count $K$ is
\begin{equation}
\Phi(z)=\prod_{i=1}^{16}\bigl[(1-p_i)+p_i z\bigr].
\end{equation}
Successive convolution and summation of coefficients from $z^k$ through $z^{16}$ give the probability of at least $k$ full scores. We calculate these probabilities separately for each template, then report averages and maxima across 140 templates, preserving differences in action spaces and scoring conditions.

\subsection{Frozen Batches and Result Traceability}
\label{app:experiment-design}
Main experiments and ablations use identical instances, images, core facts, and scoring rules, but their runs are frozen separately and paired instance by instance within each batch. Table~\ref{tab:ablation-design} defines nine ablation conditions; Table~\ref{tab:ablation-results} lists 14 prespecified comparisons. Accordingly, ablations D and E and the corresponding main-experiment Pi conditions retain their respective run scores.

Valid units determine scoring denominators. Runtime records account for infrastructure interruptions and incurred costs.

\FloatBarrier
\section{System Adaptation and Fixed Parameters}
\label{app:implementation}

\subsection{Shared Models and Context Management}
Table~\ref{tab:experiment-config} lists shared model, sampling, and hardware settings. Pi 0.99.2 (\href{https://github.com/badlogic/pi-mono/tree/005af57d88ee23b33778f343a9595b32e67ff788}{\texttt{pi-mono@005af57d}}) retains native tools, workspace, and automatic compaction. All five Pi configurations retain 4096 recent-context tokens. The 128K window reserves 65,536 tokens and triggers compaction above 65,536; the 262K window reserves 16,384 and triggers above 245,760. Summary budgets vary with the reserve, capped by the 16,384-token generation limit. Same-session and new-session parameters match within each window.

Main generation limits are 16,384 tokens for Pi and 1024 for SpaceMind. For auxiliary generation, LightMem retains its native 16000-token budget, and Mem0's JS OpenAI calls use 16,384; CoM requires none for memory construction or evidence-chain organization. All four external modules use full 2048-dimensional \href{https://huggingface.co/Qwen/Qwen3-VL-Embedding-2B}{Qwen3-VL-Embedding-2B} vectors~\cite{qwen2026vlembedding}. Queries use a fixed general conversational retrieval instruction, while writing and updating use the official default representation instruction.

\subsection{Writing, Retrieval, and Adaptation of Memory Modules}
Table~\ref{tab:memory-implementation} lists fixed module settings. Automatic retrieval and agent-initiated queries deliver memory through Pi's public interfaces, retaining each module's organization and update mechanisms.

\begin{table*}[htbp]
\setstretch{1}
\fontsize{10}{12}\selectfont
\setlength{\tabcolsep}{4pt}\renewcommand{\arraystretch}{1.15}
\caption{Memory modules integrated with Pi and their fixed parameters. All character limits count Unicode characters. Commit identifiers link to the corresponding source code.}\label{tab:memory-implementation}
\centering
\begin{tabularx}{\textwidth}{@{}C{22mm}Y Y Y@{}}
\toprule[0.9pt]
\textbf{Module and version} & \textbf{Memory writing and organization} & \textbf{Retrieval and delivery parameters} & \textbf{Adaptation and updating}\\
\midrule[0.7pt]
RAG & Single index; 1000-character text chunks with 100-character overlap; image-containing histories jointly encode text and images from the same turn & Cosine top5; one joint query with current text and images per turn; returns original historical text, images, sources, and timestamps & Supports active search; direct retrieval without additional summarization or reranking\\
\midrule[0.25pt]
\shortstack{Mem0\\JS OSS 3.3.1\\\href{https://github.com/mem0ai/mem0/tree/94c3fe9f238f3dbf29c9ce98643bd71eb13077cd}{\texttt{94c3fe9f}}} & Independent SQLite; retains native extraction, updating, and semantic memory & Retrieval limit 20; threshold 0.1 & Retains native image-description generation and text-memory pathways\\
\midrule[0.25pt]
\shortstack{CoM\\\href{https://github.com/Xiucheng-Xu/CoM/tree/52aa7ffd641059435c4585b6d9dad660518be635}{\texttt{52aa7ffd}}} & Stores public user text and ordinarily visible assistant responses by protocol turn, forming persistent raw nodes & Cosine top20; 3 anchors, product gating, 0.5 cutoff; chain context uses \texttt{average} mode; nodes are embedded using complete text & Automatic retrieval: 16384 characters; 32768 including tool search in the same turn; preserves chain order, retaining the beginning and end of long entries within a shared budget and marking omissions; raw storage remains complete\\
\midrule[0.25pt]
\shortstack{LightMem\\\href{https://github.com/zjunlp/LightMem/tree/8449d574df6bae1bdf3314a1564da65e2f37e046}{\texttt{8449d574}}} & flat LightMemory; LLMLingua-2 compression and segmentation; compression rate 0.6; sensory and short-term buffers each 512 & Retrieval top60; offline update similarity threshold 0.9; completes final-segment extraction and offline organization after history ingestion & Fixes global source ID--topic correspondence, final-segment loss, and reuse of stale vectors after updating; re-embeds updated memories and records usage\\
\bottomrule
\end{tabularx}

\end{table*}

\subsection{Native Architecture and Information Visibility}
SpaceMind uses official commit fafa4321 with native ReAct, skill routing, and test-time learning~\cite{wu2026spacemind}. Its history manager retains complete observations for eight recent steps and the first 320 normalized characters of earlier observations. At the 128K/262K windows, this step-based policy is retained, and input and output budgets are checked separately.

Modules receive public inputs, ordinarily visible assistant text, and actual environment feedback; hidden reasoning, backend scores, and newly injected retrievals are excluded from writeback. Image handling varies: RAG stores and retrieves text and images; Mem0 uses native image descriptions to form text memories; CoM and LightMem store text while Pi receives images. LightMem's invalid or cross-batch sources count as agent failures.

Source code, dependencies, model weights, container images, and adaptations are fixed and recorded by version or identifier. Main and ablation commits are \texttt{447871cf8434} and \texttt{e7bcbbfb34aa}, respectively. For main reasoning, auxiliary generation, and embedding, we separately record models, sampling settings, complete inputs, output budgets, and actual usage.

\FloatBarrier
\section{Execution Cases and Complete Results by Subcategory}
\label{app:detailed-results}

\subsection{Evidence and Scoring in Execution Cases}
Table~\ref{tab:case-studies} compares recorded new-session executions of two task instances with clean history and a 262K window, linking evidence, actions, and scores. The repair case examines evidence delivery and object matching; the parameter case compares resource costs of correct answers.

\begin{table*}[htbp]
\setstretch{1}
\fontsize{10}{12}\selectfont
\setlength{\tabcolsep}{4pt}\renewcommand{\arraystretch}{1.05}
\caption{Historical conditions, execution behavior, and scoring consequences for two actual tasks.}\label{tab:case-studies}\label{tab:spacemind-cases}
\centering
\begin{tabularx}{\textwidth}{@{}C{25mm}Y C{34mm}@{}}
\toprule[0.9pt]
\textbf{Aspect examined} & \textbf{Historical evidence and actual execution} & \textbf{Score and consequences}\\
\midrule[0.7pt]
\multicolumn{3}{C{\dimexpr\textwidth-4\tabcolsep\relax}}{\textbf{S3-O2-001: W8 on the current unit has not been test-run; W8 on another unit Q has completed its test run}}\\*
Valid evidence & H08 describes the current unit's state. The agent cites H08 and assigns removal of A and C to arms A and B, respectively, advancing currently executable operations. & 100 points\\
Object mismatch & Retrieves H08 and H16 but cites feedback H16 from unit Q and closes repair of the current unit. & 0 points\\
Missing evidence & Assigns removal of A and C without submitting valid historical evidence. & 0 points\\
\midrule[0.25pt]
\multicolumn{3}{C{\dimexpr\textwidth-4\tabcolsep\relax}}{\textbf{R3-O2-001: J01 stores successful parameter 12 for current equipment A, operation X, configuration F11, and operating condition T1}}\\*
Reuse of an applicable result & The current combination matches J01, so the agent directly submits 12; parameter 19 for equipment B does not apply to the current equipment. & Energy: 0; 100 points\\
Additional computation & Makes one computation-service request to obtain inputs and perform the calculation, then correctly submits 12. & Energy: 30; 90 points\\
\bottomrule
\end{tabularx}
\par\vspace{3pt}\noindent\begin{minipage}{\textwidth}\footnotesize
Instance identifiers omit FM. Repair rows correspond to Pi+Mem0, Pi+RAG, and Pi; direct parameter reuse is observed for both Pi+Mem0 and Pi+RAG, while the additional-computation row shows Pi. Energy is simulated computation energy within the task, initially 300 units.
\end{minipage}\end{table*}

\subsection{Main Experiment Results by Subchallenge, Scenario Task, and Difficulty}
Table~\ref{tab:main-domain-results} gives all 16 subchallenge means, separating E3 image and nonimage inputs. Table~\ref{tab:main-scenario-results} stratifies the same test set by nine tasks and three difficulty levels.

\begin{table*}[htbp]
\setstretch{1}
\fontsize{10}{12}\selectfont
\setlength{\tabcolsep}{1.8pt}\renewcommand{\arraystretch}{1.04}
\caption{Main results across 16 memory subchallenges. A valid score was recorded for every system on every instance in each category.}\label{tab:main-domain-results}
\centering
\begin{tabularx}{\textwidth}{@{}C{33mm}C{7mm}*{6}{Y}@{}}
\toprule[0.9pt]
\textbf{Category} & \textbf{$n$} & \textbf{Pi} & \textbf{RAG} & \textbf{Mem0} & \textbf{CoM} & \textbf{Light-}\newline\textbf{Mem} & \textbf{Space-}\newline\textbf{Mind$^{*}$}\\
\midrule[0.7pt]
C1 Instruction retention & 144 & \shortstack{74.69/\\16.10} & \shortstack{86.11/\\33.67} & \shortstack{93.30/\\85.58} & \shortstack{87.99/\\39.34} & \shortstack{81.78/\\28.52} & \shortstack{39.80/\\15.78}\\
C2 Rule disambiguation & 144 & \shortstack{64.58/\\22.92} & \shortstack{68.75/\\45.14} & \shortstack{71.53/\\52.78} & \shortstack{69.44/\\40.97} & \shortstack{62.50/\\27.08} & \shortstack{31.94/\\20.83}\\
C3 Downlink confirmation & 144 & \shortstack{63.35/\\6.21} & \shortstack{65.37/\\16.37} & \shortstack{68.73/\\60.80} & \shortstack{70.93/\\18.45} & \shortstack{65.46/\\7.30} & \shortstack{15.31/\\8.02}\\
T1 Validity assessment & 144 & \shortstack{76.81/\\29.56} & \shortstack{79.86/\\64.06} & \shortstack{77.38/\\61.80} & \shortstack{73.96/\\56.42} & \shortstack{77.20/\\32.06} & \shortstack{36.35/\\29.56}\\
T2 Deferred tasks & 144 & \shortstack{79.86/\\33.33} & \shortstack{84.72/\\61.81} & \shortstack{87.50/\\61.81} & \shortstack{83.33/\\54.86} & \shortstack{81.25/\\31.25} & \shortstack{53.47/\\31.25}\\
E1 Experience matching & 144 & \shortstack{87.33/\\25.12} & \shortstack{91.96/\\74.99} & \shortstack{93.06/\\88.67} & \shortstack{94.68/\\76.23} & \shortstack{87.73/\\51.25} & \shortstack{69.81/\\18.31}\\
E2 Experience revision & 144 & \shortstack{89.78/\\26.19} & \shortstack{91.15/\\58.55} & \shortstack{88.27/\\78.13} & \shortstack{89.62/\\69.64} & \shortstack{92.49/\\47.58} & \shortstack{62.85/\\26.55}\\
E3 Localization recovery & 128 & \shortstack{79.69/\\4.69} & \shortstack{89.06/\\68.75} & \shortstack{88.28/\\70.31} & \shortstack{87.50/\\65.62} & \shortstack{87.50/\\45.31} & \shortstack{31.25/\\2.34}\\
Of which: image inputs & 32 & \shortstack{53.12/\\0.00} & \shortstack{68.75/\\50.00} & \shortstack{65.62/\\43.75} & \shortstack{71.88/\\37.50} & \shortstack{65.62/\\21.88} & \shortstack{21.88/\\3.12}\\
Of which: nonimage inputs & 96 & \shortstack{88.54/\\6.25} & \shortstack{95.83/\\75.00} & \shortstack{95.83/\\79.17} & \shortstack{92.71/\\75.00} & \shortstack{94.79/\\53.12} & \shortstack{34.38/\\2.08}\\
I1 Single-operation state & 128 & \shortstack{90.23/\\20.70} & \shortstack{91.80/\\51.56} & \shortstack{84.38/\\66.41} & \shortstack{91.41/\\44.53} & \shortstack{88.67/\\25.78} & \shortstack{49.22/\\21.88}\\
I2 Multi-operation state & 112 & \shortstack{64.92/\\34.01} & \shortstack{68.12/\\39.91} & \shortstack{64.60/\\46.62} & \shortstack{69.79/\\49.94} & \shortstack{64.76/\\34.63} & \shortstack{33.95/\\31.04}\\
R1 Energy-use experience & 144 & \shortstack{78.89/\\59.77} & \shortstack{80.97/\\70.42} & \shortstack{82.81/\\73.51} & \shortstack{78.62/\\71.31} & \shortstack{82.14/\\65.04} & \shortstack{70.38/\\66.00}\\
R2 Energy reservation & 144 & \shortstack{52.94/\\20.90} & \shortstack{68.86/\\40.63} & \shortstack{59.95/\\40.28} & \shortstack{64.65/\\30.27} & \shortstack{60.86/\\26.50} & \shortstack{29.80/\\24.28}\\
R3 Result reuse & 144 & \shortstack{86.85/\\33.94} & \shortstack{88.54/\\73.23} & \shortstack{85.96/\\75.17} & \shortstack{89.93/\\73.14} & \shortstack{84.81/\\45.14} & \shortstack{47.77/\\32.44}\\
S1 Safety prerequisites & 144 & \shortstack{91.02/\\0.32} & \shortstack{94.81/\\53.56} & \shortstack{90.09/\\60.05} & \shortstack{92.08/\\58.24} & \shortstack{88.56/\\7.22} & \shortstack{41.90/\\0.00}\\
S2 Capability updates & 144 & \shortstack{84.55/\\1.39} & \shortstack{87.62/\\56.37} & \shortstack{82.29/\\54.28} & \shortstack{88.08/\\48.73} & \shortstack{87.50/\\4.86} & \shortstack{49.02/\\0.00}\\
S3 Fault recovery & 144 & \shortstack{84.24/\\2.29} & \shortstack{84.58/\\38.12} & \shortstack{86.32/\\64.79} & \shortstack{80.00/\\39.44} & \shortstack{78.82/\\2.71} & \shortstack{39.65/\\0.00}\\
\bottomrule
\end{tabularx}
\par\vspace{3pt}
\noindent\begin{minipage}{\textwidth}\fontsize{10}{12}\selectfont
Cells give same-session/new-session means; counts are per group. E3's image and nonimage rows partition its 128 instances into P1/P2 and six other applicable tasks. RAG, Mem0, CoM, and LightMem augment Pi; $^{*}$SpaceMind uses its full architecture.
\end{minipage}
\end{table*}

\begin{table*}[htbp]
\setstretch{1}
\fontsize{10}{12}\selectfont
\setlength{\tabcolsep}{1.8pt}\renewcommand{\arraystretch}{1.04}
\caption{Main results by nine scenario tasks and three difficulty levels: two stratifications of the same test set.}\label{tab:main-scenario-results}
\centering
\begin{tabularx}{\textwidth}{@{}C{33mm}C{7mm}*{6}{Y}@{}}
\toprule[0.9pt]
\textbf{Category} & \textbf{$n$} & \textbf{Pi} & \textbf{RAG} & \textbf{Mem0} & \textbf{CoM} & \textbf{Light-}\newline\textbf{Mem} & \textbf{Space-}\newline\textbf{Mind$^{*}$}\\
\midrule[0.7pt]
P1 Area surveying & 256 & \shortstack{78.35/\\17.35} & \shortstack{84.27/\\55.79} & \shortstack{78.98/\\67.14} & \shortstack{83.44/\\51.37} & \shortstack{78.26/\\23.42} & \shortstack{42.55/\\17.34}\\
P2 Sampling at designated sites & 256 & \shortstack{76.48/\\21.03} & \shortstack{82.98/\\52.89} & \shortstack{80.83/\\63.76} & \shortstack{82.71/\\48.89} & \shortstack{81.95/\\33.47} & \shortstack{41.79/\\22.64}\\
P3 Cargo transport & 256 & \shortstack{78.49/\\21.49} & \shortstack{81.74/\\50.37} & \shortstack{81.35/\\64.51} & \shortstack{82.56/\\48.68} & \shortstack{76.31/\\29.84} & \shortstack{39.77/\\19.47}\\
O1 Target inspection & 240 & \shortstack{79.01/\\20.17} & \shortstack{79.10/\\53.83} & \shortstack{81.79/\\62.95} & \shortstack{82.40/\\53.06} & \shortstack{77.79/\\27.37} & \shortstack{45.70/\\21.75}\\
O2 Equipment repair & 256 & \shortstack{79.08/\\22.59} & \shortstack{82.58/\\52.84} & \shortstack{80.41/\\66.40} & \shortstack{78.67/\\48.57} & \shortstack{78.95/\\37.46} & \shortstack{46.02/\\21.14}\\
O3 On-orbit replenishment & 256 & \shortstack{84.17/\\21.72} & \shortstack{81.87/\\53.84} & \shortstack{82.70/\\62.22} & \shortstack{83.58/\\51.82} & \shortstack{79.89/\\30.01} & \shortstack{45.34/\\19.98}\\
D1 Cruise support & 256 & \shortstack{78.71/\\23.57} & \shortstack{87.06/\\52.85} & \shortstack{83.66/\\64.72} & \shortstack{85.98/\\61.47} & \shortstack{83.54/\\36.23} & \shortstack{46.25/\\21.13}\\
D2 Scientific observation & 256 & \shortstack{75.20/\\19.18} & \shortstack{84.89/\\55.63} & \shortstack{85.79/\\68.81} & \shortstack{81.07/\\58.23} & \shortstack{82.80/\\26.74} & \shortstack{44.30/\\20.40}\\
D3 Data return & 208 & \shortstack{73.46/\\22.31} & \shortstack{79.26/\\48.40} & \shortstack{79.36/\\67.21} & \shortstack{77.32/\\47.97} & \shortstack{76.04/\\24.19} & \shortstack{45.54/\\20.65}\\
\midrule[0.25pt]
Easy & 840 & \shortstack{81.75/\\22.32} & \shortstack{84.59/\\57.61} & \shortstack{83.66/\\71.64} & \shortstack{84.10/\\55.76} & \shortstack{82.81/\\32.17} & \shortstack{51.89/\\21.71}\\
Medium & 840 & \shortstack{77.01/\\19.98} & \shortstack{83.64/\\52.23} & \shortstack{80.21/\\62.85} & \shortstack{82.44/\\50.93} & \shortstack{78.59/\\28.51} & \shortstack{41.19/\\19.08}\\
Hard & 560 & \shortstack{74.65/\\20.65} & \shortstack{78.61/\\47.36} & \shortstack{80.99/\\59.37} & \shortstack{78.47/\\49.22} & \shortstack{76.27/\\28.98} & \shortstack{36.78/\\20.75}\\
\bottomrule
\end{tabularx}
\par\vspace{3pt}
\noindent\begin{minipage}{\textwidth}\fontsize{10}{12}\selectfont
Cells give same-session/new-session means; counts are per group. RAG, Mem0, CoM, and LightMem augment Pi; $^{*}$SpaceMind uses its full architecture.
\end{minipage}
\end{table*}

\subsection{Absolute Ablation Scores and Subchallenge Profiles}
Table~\ref{tab:ablation-details} gives means, intervals, and full-score rates for nine ablation conditions, complementing the 14 paired differences in Table~\ref{tab:ablation-results}. Figure~\ref{fig:subchallenge-main} shows each group's subchallenge profile; letters follow Table~\ref{tab:ablation-design}.

\begin{table*}[htbp]
\caption{Nine ablation conditions, each with 2240/2240 first valid scores (20,160 evaluations). Means and intervals use a 0--100 scale; full-score percentages use logical full-score flags from scoring functions.}\label{tab:ablation-details}
\setstretch{1}
\centering
\fontsize{10}{12}\selectfont
\setlength{\tabcolsep}{2.5pt}\renewcommand{\arraystretch}{1.05}
\begin{tabularx}{\textwidth}{@{}C{12mm}Y C{60mm}C{25mm}@{}}
\toprule[0.9pt]
\textbf{Group} & \textbf{Configuration and condition} & \textbf{Mean score [95\% CI]} & \textbf{Full-score rate/\%}\\
\midrule[0.7pt]

A & Pi; Clean/same/262K & 83.12 [81.94,84.29] & 74.46\\
B & Pi; Clean/new/262K & 21.10 [19.92,22.30] & 11.96\\
C & Pi; Background/same/262K & 74.16 [72.72,75.56] & 65.62\\
D & Pi; Background/new/262K & 20.27 [19.11,21.42] & 11.12\\
E & Pi; Background/same/128K & 77.69 [76.36,79.07] & 69.20\\
\midrule[0.25pt]
F & Pi+Mem0; Clean/same/262K & 87.56 [86.42,88.66] & 81.65\\
G & Pi+Mem0; Clean/new/262K & 75.71 [74.31,77.09] & 67.46\\
\midrule[0.25pt]
H & Pi+RAG; Clean/same/262K & 88.63 [87.55,89.66] & 82.59\\
I & Pi+RAG; Clean/new/262K & 59.43 [57.86,60.98] & 50.36\\
\bottomrule[0.9pt]
\end{tabularx}

\par\vspace{3pt}
\noindent\begin{minipage}{\textwidth}\fontsize{10}{12}\selectfont
Same/new denote same-session/new-session; background denotes routine operational history. Windows of 128K and 262K contain 131,072 and 262,144 tokens.
The 95\% intervals synchronously resample 2089 construction groups within 140 equally weighted templates: 10,000 resamples, seed 1729.
\end{minipage}
\end{table*}

{\setlength{\intextsep}{4pt}\setlength{\abovecaptionskip}{5pt}
\begin{figure*}[tbp]
\centering\includegraphics[width=\textwidth]{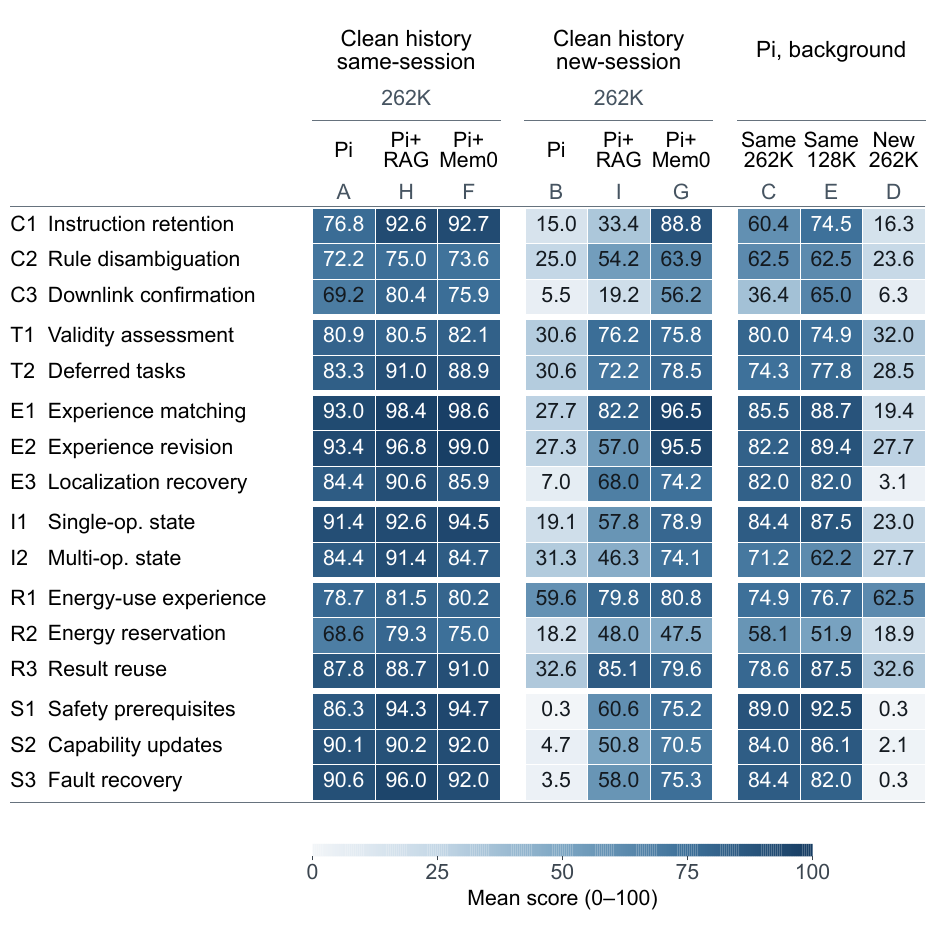}
\caption{Subchallenge means for nine ablation conditions. Left and middle: clean-history same-session and new-session evaluation at 262K. Right: Pi with background, with windows and sessions labeled; E triggers native compaction. Letters follow Table~\ref{tab:ablation-design}. Each group has 2240 instances: 128 each for E3 and I1, 112 for I2, and 144 for each remaining subchallenge.}
\label{fig:subchallenge-main}
\end{figure*}
}

\FloatBarrier
\end{document}